\documentclass[journal]{IEEEtran}
\usepackage{times}
\usepackage{epsfig}
\usepackage{graphicx}
\usepackage{amsmath}
\usepackage{amssymb}
\usepackage{array}
\usepackage{subcaption}
\usepackage{color}
\usepackage{multirow}

\usepackage{fancyhdr}
\fancypagestyle{firstpage}{%
  \rhead{AB-ReLU}
  \lfoot{\small{This paper is published by Multimedia Tools and Applications, Springer, available at: https://doi.org/10.1007/s11042-020-10269-x. \\ The copyright has been transferred to Springer.}}
}

\begin{document}

\title{Average Biased ReLU Based CNN Descriptor for Improved Face Retrieval}

\author{Shiv Ram Dubey, Soumendu Chakraborty
\thanks{S.R. Dubey is with the Computer Vision Group, Indian Institute of Information Technology (IIIT), Sri City, Andhra Pradesh, India (Email: srdubey@iiits.in).}
\thanks{S. Chakraborty is with the Indian Institute of Information Technology (IIIT), Lucknow, Uttar Pradesh, India (Email: soum.uit@gmail.com ).
}
}

\maketitle
\thispagestyle{firstpage}

\begin{abstract}
The convolutional neural networks (CNN), including AlexNet, GoogleNet, VGGNet, etc. extract features for many computer vision problems which are very discriminative. The trained CNN model over one dataset performs reasonably well whereas on another dataset of similar type the hand-designed feature descriptor outperforms the same trained CNN model. The Rectified Linear Unit (ReLU) layer discards some values in order to introduce the non-linearity. In this paper, it is proposed that the discriminative ability of deep image representation using trained model can be improved by Average Biased ReLU (AB-ReLU) at the last few layers. Basically, AB-ReLU improves the discriminative ability in two ways: 1) it exploits some of the discriminative and discarded negative information of ReLU and 2) it also neglects the irrelevant and positive information used in ReLU. The VGGFace model trained in MatConvNet over the VGG-Face dataset is used as the feature descriptor for face retrieval over other face datasets. The proposed approach is tested over six challenging, unconstrained and robust face datasets (PubFig, LFW, PaSC, AR, FERET and ExtYale) and also on a large scale face dataset (PolyUNIR) in retrieval framework. It is observed that the AB-ReLU outperforms the ReLU when used with a pre-trained VGGFace model over the face datasets. The validation error by training the network after replacing all ReLUs with AB-ReLUs is also observed to be favorable over each dataset. The AB-ReLU even outperforms the state-of-the-art activation functions, such as Sigmoid, ReLU, Leaky ReLU and Flexible ReLU over all seven face datasets.
\end{abstract}

\section{Introduction}
The image descriptors are generally used to perform the image matching. Most of the research in the early days was focused on designing the hand-crafted descriptors which have shown promising performance in several computer vision applications, such as image matching \cite{iold}, identity verification \cite{sharma2015identity}, face recognition \cite{lbpface,lghp}, image retrieval \cite{mdlbp}, biomedical image analysis \cite{lwp,lbdp,ldep}, etc. 
Several hand-designed descriptors have been proposed for face retrieval as well \cite{lghp,ldop,ldgp,csqp,ldrp,fdlbp}. However, the discriminative power of these hand-crafted descriptors is limited as they are not able to utilize the context of data.

Since last few years, the convolutional neural networks (CNN) have attracted the full attention of researchers in computer vision community. 
The AlexNet \cite{alexnet} was the first CNN model that won the Imagenet classification challenge \cite{imagenet} in 2012 with a great margin as compared to the best performing hand-designed features. After Alexnet, several CNN models have been proposed for the Imagenet classification, such as VGGNet \cite{vggnet}, GoogLeNet \cite{googlenet} and ResNet \cite{resnet}. The network became deeper over time , i.e., AlexNet (8 stages) in 2012, VGGNet (16 and 19 stages) in 2014, GoogLeNet (22 stages) in 2012, and ResNet (152 stages) in 2015. The CNNs have been also used for different applications, such as multimedia classification \cite{shamsolmoali2019high}, image denoising \cite{wang2019multi}, and image retrieval \cite{li2019piecewise}.

The deep neural network based models have also been proposed for the face recognition task. Some recent and renowned deep learning based approaches are DeepFace \cite{deepface}, FaceNet \cite{facenet}, and VGGFace \cite{vggface} amongst others for face recognition. The DeepFace model has used 9 learnable layers for face representation \cite{deepface}. The number of parameters in the DeepFace model is too high as it does not use the weight sharing. The DeepFace model has reported an accuracy of 97.35\% on the Labeled Faces in the Wild (LFW) dataset \cite{deepface,lfw}. FaceNet is also proposed as the feature extractor for face recognition and clustering \cite{facenet}. The FaceNet model uses the deep convolutional network for the feature embedding. The FaceNet model has reported 99.63\% accuracy over LFW face dataset. The VGGFace model has utilized the convolutional neural network (CNN) based end-to-end learning for face recognition \cite{vggface}. It is trained over a very large scale VGGFace database with 2.6M images from 2.6K subjects. In our work, we use the VGGFace model as a feature extractor for the face retrieval experiments.

The pre-trained models are also used for several tasks in the Computer Vision. 
The CNN features are fused with the hand-designed features for content-based image retrieval in \cite{liu2017fusion}.
Very recently, it has been claimed that the trained network used over still face images can also be used for face verification in the videos effectively \cite{bansal2017s}. 
Deep features are also used for image quality assessment \cite{ma2019multimedia}. A pre-trained CNN model is also used for the content based image retrieval \cite{wan2014deep}. Some researchers also used the transfer learning to utilize a trained network meant for a domain in some other domain, such as Deep transfer \cite{tzeng2015simultaneous}, etc. 
The pre-trained VGG network has also been used for remote-sensing image retrieval \cite{geexploiting}. In this paper also, the pre-trained network equipped with the proposed ABReLU is used for the face retrieval task. 

Some researchers have also focusing over different layers of the CNN model. A center loss function is used in \cite{wen2016discriminative} instead of the softmax loss function for face recognition. The ReLU discards the negative values which actually shows the absence of certain visual features and might be useful to improve the discriminative ability \cite{relu}. 
In order to get rid of negative values in ReLU, a Rectified Factor Network is introduced in \cite{clevert2015rectified}. 
A Parametric Rectified Linear Unit (PReLU) is used by He et al. It is a generalization of the Rectified Linear Unit (ReLU) where the slope of the negative region is considered as one of the parameters of each neuron \cite{he2015delving}. The ReLU also has the ``dying Gradient" problem where the gradient flowing through a unit can be zero forever \cite{alexnet}. Leaky ReLU (LReLU) tried to fix the dying gradient problem of ReLU during training by computing a small slope for the negative inputs \cite{lrelu}. The LReLU is extended to randomized leaky rectified linear units (RReLU) by introducing a random small slope \cite{rrelu}. Flexible ReLU (FReLU) is an another attempt to use the negative values based on a threshold \cite{frelu}. An exponential linear unit (ELU) is proposed in \cite{elu} which also considers the ReLU's negative values. 
One of the recent developments in this area is the advent of non-static activation functions \cite{zhou2019deep}.
Most of the existing rectifier units do not consider the negative values which might be important. These rectifier units are also not dependent upon the input data. In this paper, a new data dependent rectifier unit is proposed to boost the discriminative power of VGGFace descriptor.

The main contributions of the this paper are as follows:
\begin{itemize}
	\item The suitability of the use of pre-trained CNN model over other databases of similar type is explored.
    \item A new data dependent Average Biased Rectified Liner Unit (AB-ReLU) is proposed to boost the discriminative power of the pre-trained network.
    \item The suitability of proposed AB-ReLU is tested at different layers of the network.
    \item The image retrieval experiments are conducted over seven challenging face datasets, including a large scale face dataset.
    \item The effect of the distances is analyzed with VGGFace feature.
    \item The performance of AB-ReLU is also analyzed by training the network after replacing all ReLUs with AB-ReLUs in VGGFace model. 
\end{itemize}

One of the underlying problems of ReLU is the diminishing gradient in the negative direction for the negative inputs. In neural network training, the data are always expected to be zero-centric having unit variance. Due to the dying ReLU problem, the variance of activations becomes too low after some layers. The AB-ReLU is designed to encounter this problem up to some extent by allowing even negative data to be passed to the next layer. Moreover, it captures the positive or negative dominancy in the activations and tries to put through roughly half of the activations to the next layer. The intuition behind AB-ReLU can be seen from the following angle of point of view also: The initial layers in CNN capture the coarse information, hence suppressing anything that is less than zero cause’s a significant information loss. Whereas, higher layers extract abstract information which may be redundant also, so introducing a highly non-linear activation function increases the classification accuracy. The proposed AB-ReLU is able to capture the coarse and abstract information effectively.

The rest of the paper is organized as follows: Section 2 reviews the VGGFace model and rectified linear unit; Section 3 proposes a new data dependent rectified linear unit and modified VGGFace descriptor; Section 4 presents the experimental setup; Section 5 presents the results and discussions; and finally Section 6 sets the concluding remarks.

\section{Related Works}
In this section, first the original VGGFace model used in this work is described in detail and then the original rectified linear unit is presented.

\begin{table}
\caption{VGGFace Layer Description. In `Filter' column, $f$, $s$ and $p$ represent the filter size, stride and padding, respectively. In `Volume Size' column, the first value is a dimension of volume and the second value is depth of volume, i.e. 224,3 represents volume size 224$\times$224$\times$3. The last fully connected layer and softmax layer are not shown because the output of `relu7' is considered as the 4096-dimensional feature vector in this work.}
\label{table1}
\begin{center}
\begin{tabular}{|m{.4cm}|m{1cm}|m{0.7cm}|m{3.28cm}|m{1cm}|}
\hline
No. & Layer Name & Layer Type & Filter & Volume Size \\
\hline\hline
0 & input & Image & n/a & 224,3 \\
1 & conv1\_1 & Conv & $f$:3,3,64, $s$:1, $p$:1 & 224,64 \\
2 & relu1\_1 & Relu & n/a & 224,64 \\
3 & conv1\_2 & Conv & $f$:3,64,64, $s$:1, $p$:1 & 224,64 \\
4 & relu1\_2 & Relu & n/a & 224,64 \\
5 & pool1   & Pool & $f$:2, $s$:2, $p$:0 & 112,64 \\
6 & conv2\_1 & Conv & $f$:3,64,128, $s$:1, $p$:1 & 112,128 \\
7 & relu2\_1 & Relu & n/a & 112,128 \\
8 & conv2\_2 & Conv & $f$:3,128,128, $s$:1, $p$:1 & 112,128 \\
9 & relu2\_2 & Relu & n/a & 112,128 \\
10 & pool2   & Pool & $f$:2, $s$:2, $p$:0 & 56,128 \\
11 & conv3\_1 & Conv & $f$:3,128,256, $s$:1, $p$:1 & 56,256 \\
12 & relu3\_1 & Relu & n/a & 56,256 \\
13 & conv3\_2 & Conv & $f$:3,256,256, $s$:1, $p$:1 & 56,256 \\
14 & relu3\_2 & Relu & n/a & 56,256 \\
15 & conv3\_3 & Conv & $f$:3,256,256, $s$:1, $p$:1 & 56,256 \\
16 & relu3\_3 & Relu & n/a & 56,256 \\
17 & pool3   & Pool & $f$:2, $s$:2, $p$:0 & 28,256 \\
18 & conv4\_1 & Conv & $f$:3,256,512, $s$:1, $p$:1 & 28,512 \\
19 & relu4\_1 & Relu & n/a & 28,512 \\
20 & conv4\_2 & Conv & $f$:3,512,512, $s$:1, $p$:1 & 28,512 \\
21 & relu4\_2 & Relu & n/a & 28,512 \\
22 & conv4\_3 & Conv & $f$:3,512,512, $s$:1, $p$:1 & 28,512 \\
23 & relu4\_3 & Relu & n/a & 28,512 \\
24 & pool4   & Pool & $f$:2, $s$:2, $p$:0 & 14,512 \\
25 & conv5\_1 & Conv & $f$:3,512,512, $s$:1, $p$:1 & 14,512 \\
26 & relu5\_1 & Relu & n/a & 14,512 \\
27 & conv5\_2 & Conv & $f$:3,512,512, $s$:1, $p$:1 & 14,512 \\
28 & relu5\_2 & Relu & n/a & 14,512 \\
29 & conv5\_3 & Conv & $f$:3,512,512, $s$:1, $p$:1 & 14,512 \\
30 & relu5\_3 & Relu & n/a & 14,512 \\
31 & pool5   & Pool & $f$:2, $s$:2, $p$:0 & 7,512 \\
32 & fc6 & Conv & $f$:7,512,4096, $s$:1, $p$:0 & 1,4096 \\
33 & relu6 & Relu & n/a & 1,4096 \\
34 & fc7 & Conv & $f$:1,4096,4096, $s$:1, $p$:0 & 1,4096 \\
35 & relu7 & Relu & n/a & 1,4096 \\
\hline
\end{tabular}
\end{center}
\end{table}

\subsection{VGGFace Model}
In this work, the original pre-trained VGGFace model is taken from the MatConvNet library \cite{matconvnet} released by University of Oxford\footnote {$http://www.robots.ox.ac.uk/\hbox{$\scriptstyle\mathtt{\sim}$}vgg/software/vgg\_face/$}. This model is based on the implementation of VGG-Very-Deep-16 CNN architecture as described in \cite{vggface}. This model is trained over VGGFace database\footnote {$http://www.robots.ox.ac.uk/\hbox{$\scriptstyle\mathtt{\sim}$}vgg/data/vgg\_face/$} which consists 2.6M face images from 2,622 subjects. The layers of VGGFace model are summarized in Table \ref{table1}. In this table, the last fully connected layer and sofmax layer of VGGFace are not listed as they are not required in this work. The output of `relu7' is considered as the VGGFace feature descriptor. The filter size, stride and padding are mentioned in the `Filter' column with fields $f$, $s$ and $p$ respectively. A filter size $f$:3,128,256 means total 256 filters of dimension 3$\times$3 and depth 128. Similarly, a volume size 112,64 means a 3-D volume of dimension 112$\times$112 with depth 64. In this work, changes are made in selected rectified linear unit (ReLU) layers, especially in the last few layers as described in the next section.

\subsection{Rectified Linear Unit}
\begin{figure}[t]
    \centering
    \includegraphics[width=.50\linewidth]{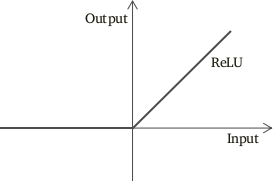}
  \caption{The original rectified linear unit (ReLU) function \cite{relu}. All the \textit{Negative} ($-ive$) input values are converted to zero, whereas all the \textit{Positive} ($+ive$) input values are passed as the identity function.}
  \label{fig:relu}
\end{figure}

The rectified linear unit (ReLU) in a neural network is used to introduce the non-linearity \cite{relu}, \cite{alexnet}. The ReLU simply works like a filter, ignores the negative signals and pass the positive signals. Consider $I_i^n$ is the input volume to ReLU at $n^{th}$ layer of any network and $I_o^{n}$ is the output volume of ReLU for $(n)^{th}$ layer. Suppose the input volume $I_i^n$ is of dimension $d$ and $D_k$ is the size of the input volume in the $k^{th}$ dimension $\forall k \in [1, d]$. Then, an element at position $\rho=(\rho_1,\rho_2,\cdots,\rho_d)$ of output volume $I_o^{n}$ is computed from the corresponding element of input volume $I_i^{n}$ as follows,
\begin{equation}
	I_o^{n}(\rho)=
    \begin{cases}
    	I_i^{n}(\rho),& \text{if } I_i^{n}(\rho) > 0\\
    	0,              & \text{otherwise}
	\end{cases}
\end{equation}
where $\rho$ is $d$-dimensional, $D_k$ is the size of $I_i^n$ in $k^{th}$ dimension, $\rho_k \in [1,D_k]$ $\forall k \in [1,d]$. 
The ReLU function is illustrated in Fig. \ref{fig:relu}. It is linear in the \textit{Positive} ($+ive$) range, whereas zero in the \textit{Negative} ($-ive$) range. The main drawback with ReLU is that it passes all $+ive$ values even they are not important and blocks all $-ive$ values irrespective of their important. This problem is solved by introducing a data dependent average biased ReLU as elaborated in the next section.

\section{Proposed Face Descriptor}
In this section, first a data dependent average biased rectified linear unit (AB-ReLU) is proposed, then it is applied to existing pre-trained VGGFace model \cite{vggface} to create a more discriminative face descriptor, and finally AB-ReLU based VGGFace descriptor is used for face retrieval.

\subsection{Average Biased Rectified Linear Unit}
It can be noticed in ReLU that it is not data dependent. It rejects all the $-ive$ signals and passes all the $+ive$ signals which can lead to less discriminative feature extraction. In this section, the problem is resolved by introducing a new data dependent ReLU named average biased rectified linear unit (AB-ReLU). The AB-ReLU is data dependent by exploiting the average property of the input volume. It also works like a filter and passes only those signals which satisfy the average biased criteria. The average biased criteria ensures that only important features get passed irrespective of its sign. Suppose, AB-ReLU is used in any network at $n^{th}$ layer and $I_i^n$ and $I_o^{n}$ are input volume and output volume for this layer respectively. Then, the $\rho^{th}$ element of output layer $I_o^{n}$ is given by following equation,
\begin{equation}
	I_o^{n}(\rho)=
    \begin{cases}
    	I_i^{n}(\rho)-\beta,& \text{if } I_i^{n}(\rho)-\beta > 0\\
    	0,              & \text{otherwise}
	\end{cases}
\end{equation}
where $\rho=(\rho_1,\rho_2,\cdots,\rho_d)$ represents the position of an element, $d$ is the dimension of $I_i^n$, $D_k$ is the size of $I_i^n$ in $k^{th}$ dimension, $\rho_k \in [1,D_k]$ $\forall k \in [1,d]$, and $\beta$ is the average biased factor defined as follows,
\begin{equation}
	\beta=\alpha \times A_i^n
\end{equation}
where $\alpha$ is a parameter to be set empirically and $A_i^n$ is the average of input volume computed as follows,
\begin{equation}
	A_i^n = \frac{\sum_{\rho_1=1}^{D_1}{\sum_{\rho_2=1}^{D_2} \cdots \sum_{\rho_d=1}^{D_d} {I_i^{n}(\rho_1,\rho_2,\cdots,\rho_d)}}}{D_1 \times D_2 \times \cdots \times D_d}
\end{equation}

\begin{figure*}[t]
\centering
  \begin{subfigure}{.45\textwidth}
    \centering
    \includegraphics[width=1\linewidth]{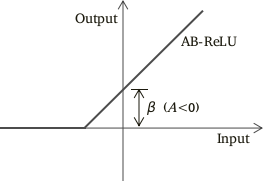}
    \caption{AB-ReLU if $A_i^n < 0$}
    \label{fig:abrelu1}
  \end{subfigure}%
  \begin{subfigure}{.45\textwidth}
    \centering
    \includegraphics[width=1\linewidth]{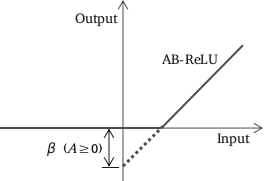}
    \caption{AB-ReLU if $A_i^n \geq 0$}
    \label{fig:abrelu2}
  \end{subfigure}
  \caption{The average biased rectified linear unit (AB-ReLU) function. Here, $\beta$ represents an average biased factor. The effective biased is $+ive$ in (a) because the value of $\beta$ is $-ive$, whereas effective biased is $-ive$ in (b) because the value of $\beta$ is $+ive$. The $-ive$ $\beta$ represents that the $-ive$ values are also important, whereas the $+ive$ $\beta$ represents that all the $+ive$ values are not important.}
  \label{fig:abrelu}
\end{figure*}

The AB-ReLU leads to two AB-ReLUs, i.e. $+ive$ AB-ReLU and $-ive$ AB-ReLU based upon the input data. This behavior of AB-ReLU is illustrated in Fig. \ref{fig:abrelu} where Fig. \ref{fig:abrelu1} shows the $+ive$ ReLU function and Fig. \ref{fig:abrelu2} depicts the $-ive$ AB-ReLU function. The $+ive$ AB-ReLU signifies the $+ive$ average biased scenario when the input data volume $A_i^n$ has the $-ive$ majority, i.e. $A_i^n < 0$ and allows some prominent $-ive$ signals by converting them into $+ive$ signal with the addition of an average biased factor of input volume ($\beta$). Similarly, if the input data volume $A_i^n$ has the $+ive$ majority, i.e. $A_i^n \geq 0$ then AB-ReLU blocks even some inferior $+ive$ signals along with all $-ive$ signals by subtracting the average biased factor of input volume ($\beta$).
The default value of $\alpha$ is set to $1$. 

The average bias computed by the AB-ReLU works like a controller to control the dominance of the positive and the negative values. If the negative values are dominant, then there is a higher possibility that these values are utilized by the AB-ReLU, as opposed to the ReLU which discards such values. Even though AB-ReLU considers the negative values, it discards the higher negatives. In case of positive dominance of the features there is a possibility of having the irrelevant lower-range positive values. Such values are discarded by the average bias of the AB-ReLU. Whereas, the ReLU does not filter the irrelevant positive features.
The ABReLU does not simply discard the negative values unlike ReLU. Rather, it finds the important negative values and adds the bias to shift it towards the positive values. The utilization of the positive feature values is not only specific to the face retrieval task. Generally, the positive values are given more importance in CNN as it is generated by convolving the input with the kernels. A kernel produces the positive values for a neuron if such types of features are present in the input, otherwise, it produces a negative value. Thus, the positive values in CNN can represent the existence of the important visual cues in the image. Finer visual information might not be covered through the learnt positive values. However, such information can play a crucial role in face related applications where the texture information in the faces is generally similar. In this paper, we utilize such features by shifting it towards the positive features.
In the next subsection, the AB-ReLU is used to construct the descriptor.

\subsection{AB-ReLU based VGGFace Descriptor}
In this subsection, the VGGFace model is used with AB-ReLU to construct the improved VGGFace descriptors. The description of the different models is summarized in Table \ref{tab:notation}. The AB-ReLU is applied directly over pre-trained VGGFace model at some layers instead of simple ReLU. The output of layer35 (i.e. ReLU) of original pre-trained VGGFace model after reshaping into a 1-D array is used as the VGGFace descriptor and represented by VGGFace35ReLU (or just 35R as shorthand notation). The first descriptor is proposed by simply replacing the last ReLU, i.e. at layer35 with AB-ReLU and converting its output into a 1-D array. This descriptor is represented by VGGFace35AB-ReLU (i.e. 35AR) for $\alpha=1$. The other variants of this descriptor are VGGFace35AB-ReLU2 (i.e. 35AR2) and VGGFace35AB-ReLU5 (i.e. 35AR5) for $\alpha=2$ and $\alpha=5$ respectively. Similarly, other descriptors are generated by replacing some ReLU of VGGFace with AB-ReLU. In second descriptor i.e. VGGFace33AB-ReLU (i.e. 33AR) for $\alpha=1$, layer34 and layer35 are removed, the ReLU at layer33 is replaced with AB-ReLU and the output of layer33 is considered as the descriptor after reshaping into a 1-D array. Its other variants are VGGFace33AB-ReLU2 (i.e. 33AR2) and VGGFace33AB-ReLU5 (i.e. 33AR5) for $\alpha=2$ and $\alpha=5$ respectively. In VGGFace33AB-ReLU\_35 (i.e. 33AR\_35) descriptor, the ReLU at layer33 is replaced with AB-ReLU while the output of layer35 using ReLU is considered as the descriptor. AB-ReLU is applied in multiple layers, i.e. at layer33 and layer35 in VGGFace33,35AB-ReLU (i.e. 33,35AR). The AB-ReLU is also applied at layer30. Two descriptors namely VGGFace30AB-ReLU (i.e. 30AR) and VGGFace30AB-ReLU\_35 (i.e. 30AR\_35) are considered for the experiments. In VGGFace30AB-ReLU, the output layer30 (i.e. AB-ReLU) is taken as the descriptor, whereas in VGGFace30AB-ReLU\_35, the AB-ReLU is used at layer30 and the output of last layer (i.e. layer35) is taken as the descriptor. In experiment section, the shorthand notations of descriptor are used. Note that all the descriptors are normalized to unit sum by dividing all the features with sum of features.

\begin{table*}[!t]
    \caption{The summary of the different models including name, short name and description.}
    \centering
    \begin{tabular}{|m{0.35\columnwidth}|m{0.12\columnwidth}|m{1.3\columnwidth}|}
         \hline
         Name & Short Name & Description \\\hline
         VGGFace35ReLU & 35R & The output of layer35 (i.e. ReLU) of original pre-trained VGGFace model.\\\hline 
         VGGFace35AB-ReLU & 35AR & The output of layer35 after replacing the ReLU at layer35 with AB-ReLU having $\alpha=1$.\\\hline
         VGGFace35AB-ReLU2 & 35AR2 & The output of layer35 after replacing the ReLU at layer35 with AB-ReLU having $\alpha=2$.\\\hline
         VGGFace35AB-ReLU5 & 35AR5 & The output of layer35 after replacing the ReLU at layer35 with AB-ReLU having $\alpha=5$.\\\hline
         VGGFace33ReLU & 33R & The output of layer33 (i.e. ReLU) of original pre-trained VGGFace model.\\\hline 
         VGGFace33AB-ReLU & 33AR & The output of layer33 after replacing the ReLU at layer33 with AB-ReLU having $\alpha=1$.\\\hline
         VGGFace33AB-ReLU2 & 33AR2 & The output of layer33 after replacing the ReLU at layer33 with AB-ReLU having $\alpha=2$.\\\hline
         VGGFace33AB-ReLU5 & 33AR5 & The output of layer33 after replacing the ReLU at layer33 with AB-ReLU having $\alpha=5$.\\\hline
         VGGFace33AB-ReLU\_35 & 33AR\_35 & The  ReLU  at  layer33  is  replaced with  AB-ReLU  while  the  output  of  layer35  using  ReLU  is  considered  as  the  descriptor.\\\hline
         VGGFace33,35AB-ReLU & 33,35AR & The output of layer35 after applying AB-ReLU at layer33 and layer35. \\\hline
         VGGFace30ReLU & 30R & The output of layer30 (i.e. ReLU) of original pre-trained VGGFace model.\\\hline 
         VGGFace30AB-ReLU & 30AR & The output of layer30 after replacing the ReLU at layer30 with AB-ReLU having $\alpha=1$.\\\hline
         VGGFace30AB-ReLU\_35 & 30AR\_35 & The  ReLU  at  layer30  is  replaced with the AB-ReLU  while  the  output  of  layer35  using  ReLU  is  considered  as  the  descriptor. \\\hline
    \end{tabular}
    \label{tab:notation}
\end{table*}

\begin{figure*}
  \hspace{0.5em}
  \begin{subfigure}{.2\textwidth}
    \centering
    \includegraphics[width=1\linewidth]{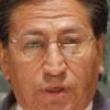}
    \caption{Example Image}
    \label{fig:eximage}
  \end{subfigure}%
  \hspace{0.5em}
  \begin{subfigure}{.33\textwidth}
    \centering
    \includegraphics[width=1\linewidth]{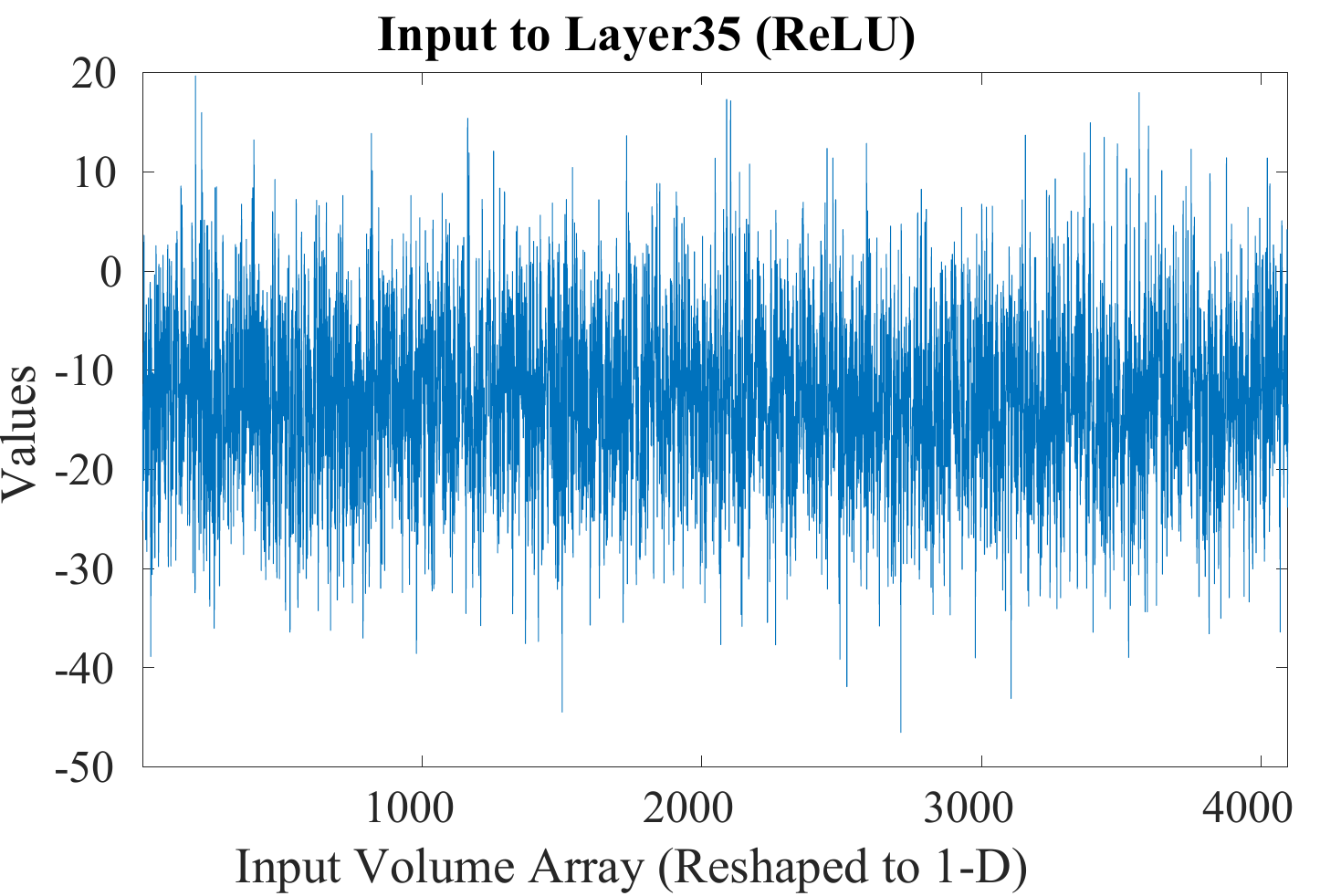}
    \caption{Input to Layer 35}
    \label{fig:exinput}
  \end{subfigure}%
  \begin{subfigure}{.33\textwidth}
    \centering
    \includegraphics[width=1\linewidth]{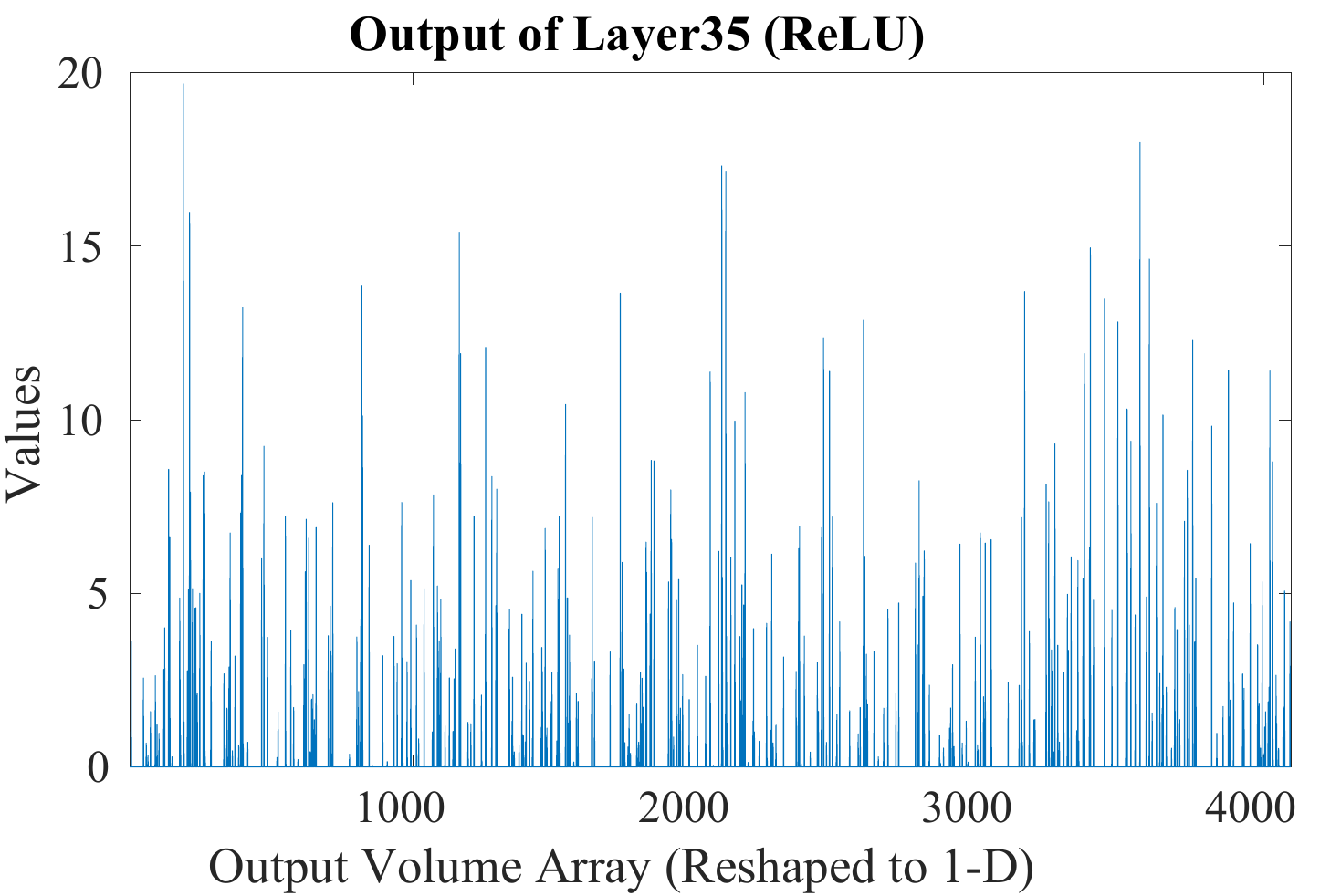}
    \caption{Layer 35 (ReLU)}
    \label{fig:exoutput-relu}
  \end{subfigure}
  \begin{subfigure}{.33\textwidth}
    \centering
    \includegraphics[width=1\linewidth]{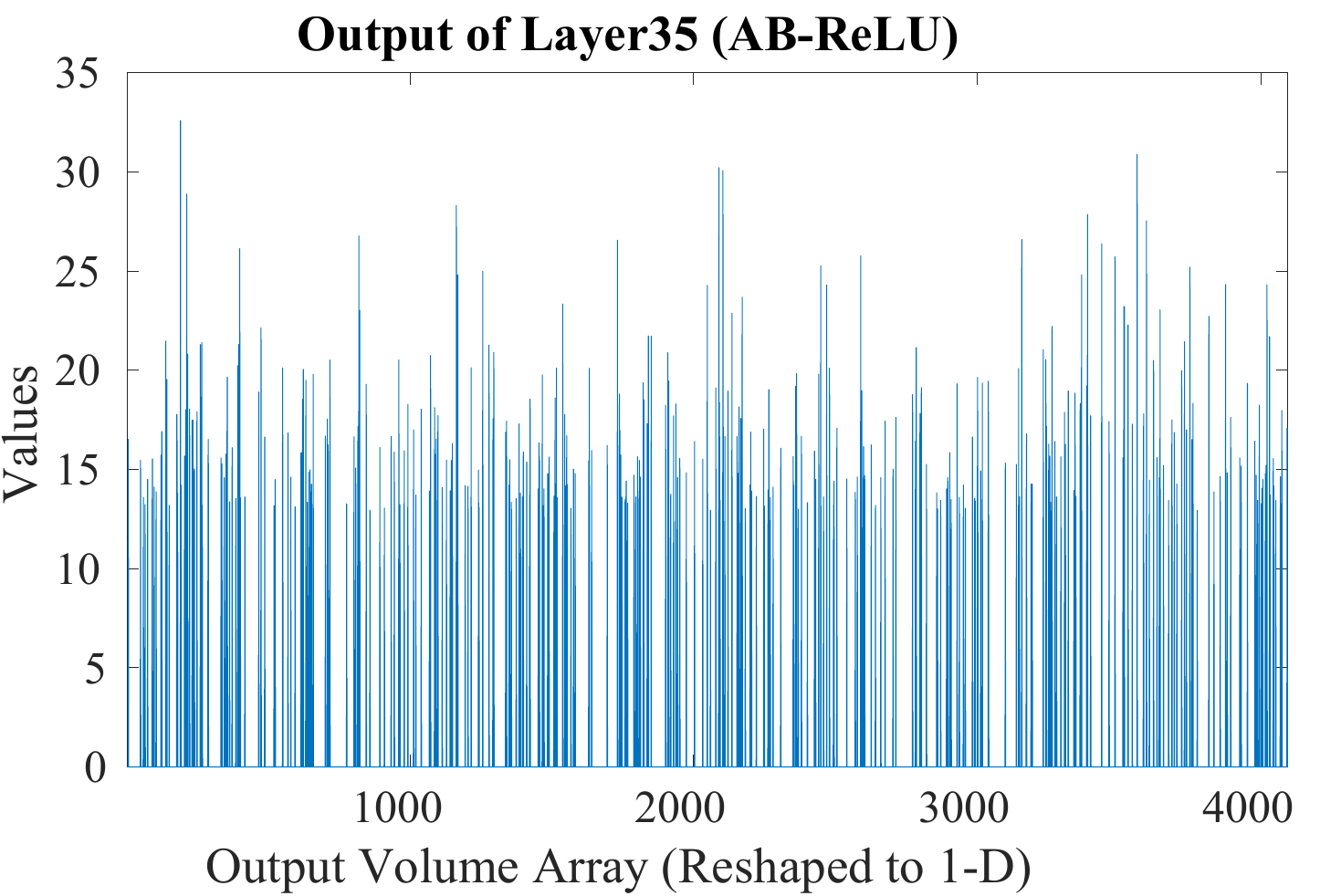}
    \caption{Layer 35 (AB-ReLU, $\alpha=1$)}
    \label{fig:exoutput-abrelu}
  \end{subfigure}%
  \hspace{0.5em}
  \begin{subfigure}{.33\textwidth}
    \centering
    \includegraphics[width=1\linewidth]{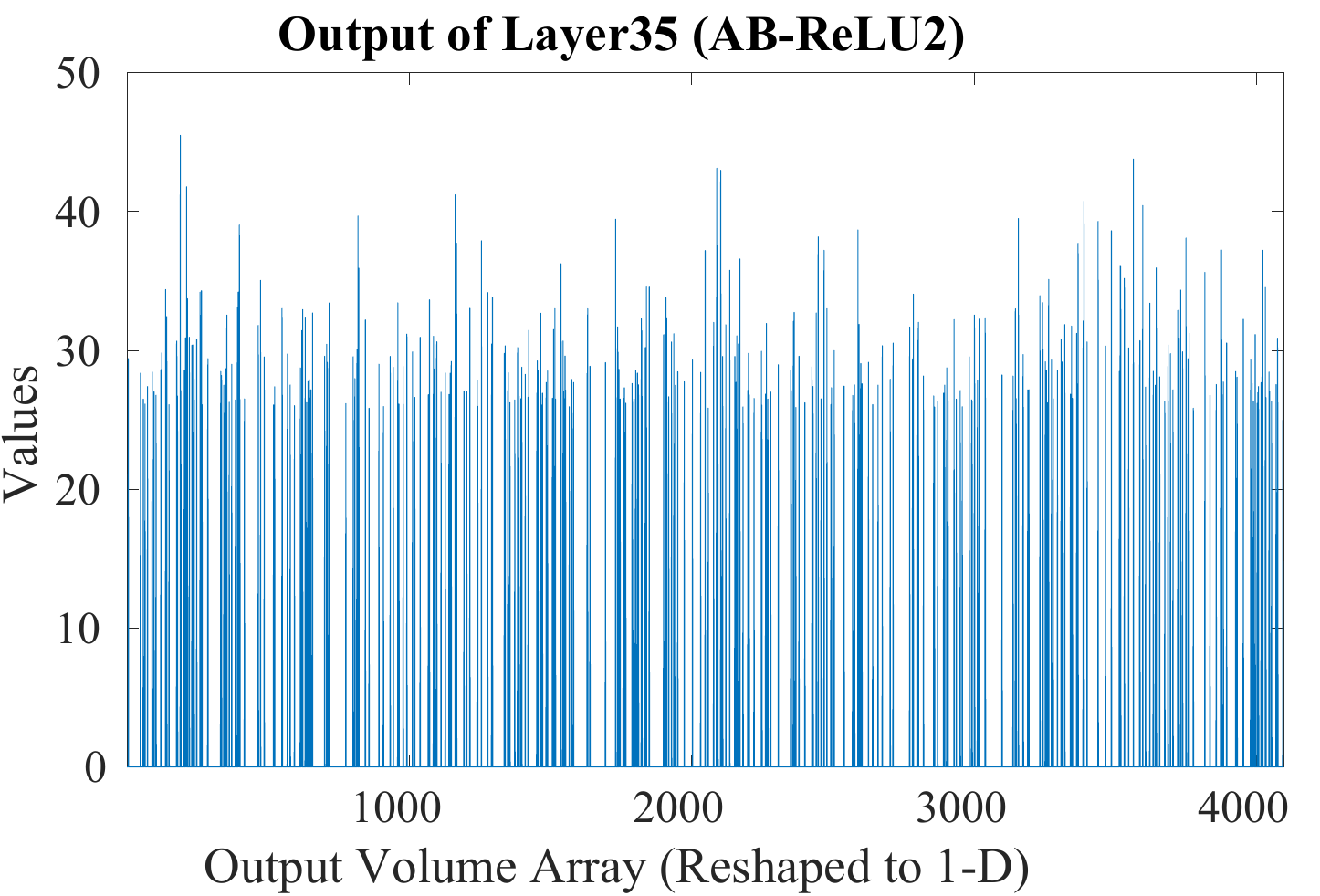}
    \caption{Layer 35 (AB-ReLU, $\alpha=2$)}
    \label{fig:exoutput-abrelu2}
  \end{subfigure}%
  \begin{subfigure}{.33\textwidth}
    \centering
    \includegraphics[width=1\linewidth]{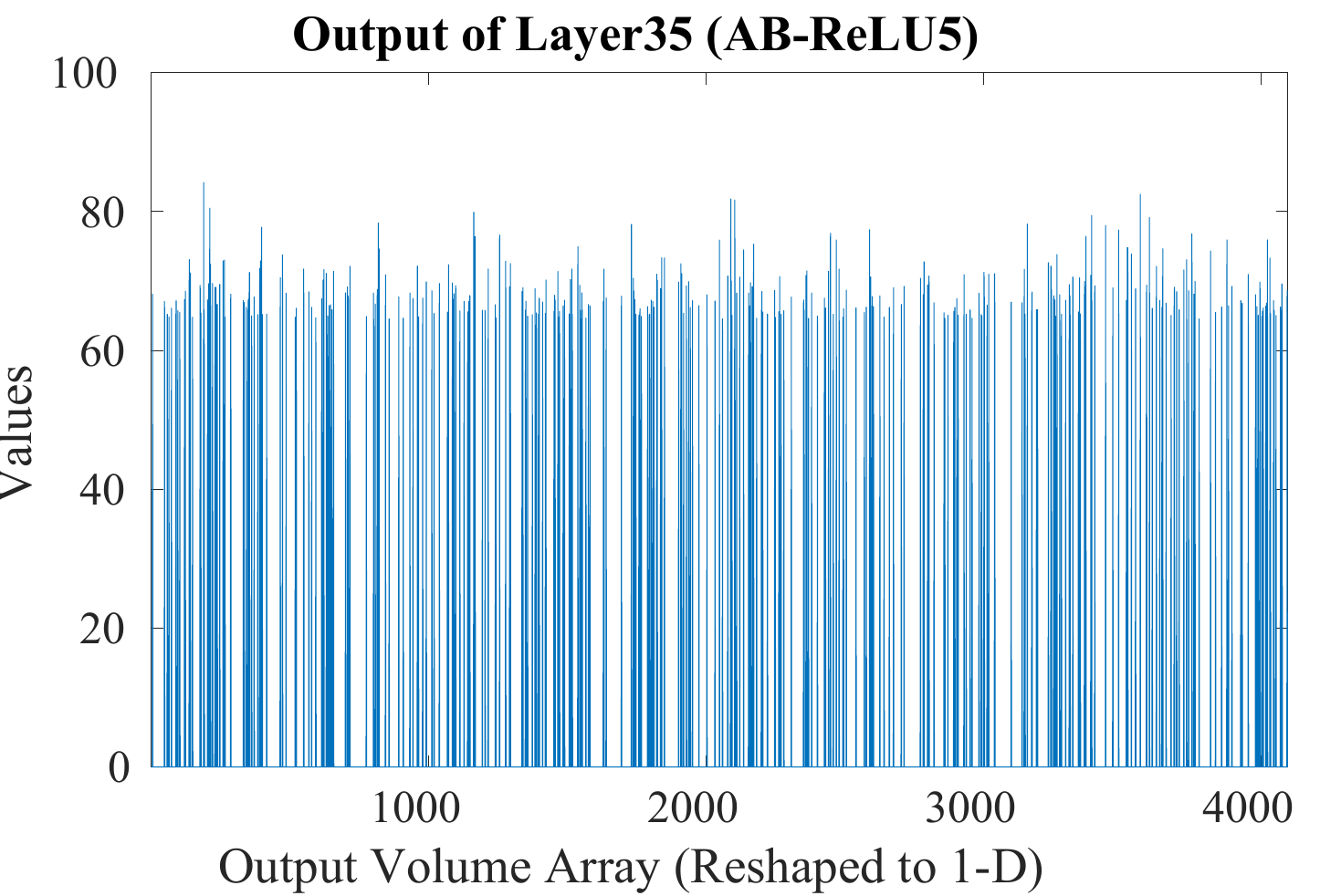}
    \caption{Layer 35 (AB-ReLU, $\alpha=5$)}
    \label{fig:exoutput-abrelu5}
  \end{subfigure}
  \caption{An example illustrating the ReLU and AB-ReLU in terms of the final feature of layer 35 of VGGFace model. (a) An Example Image from LFW database \cite{lfw}. (b) Input to Layer 35. (c) Output of Layer 35 (ReLU). (d) Output of Layer 35 (AB-ReLU, $\alpha=1$). (e) Output of Layer 35 (AB-ReLU, $\alpha=2$). (f) Output of Layer 35 (AB-ReLU, $\alpha=5$). }
  \label{fig:example}
\end{figure*}

The effect of AB-ReLU with pre-trained VGGFace (VGGFace35AB-ReLU) is illustrated with an example face image in Fig. \ref{fig:example}. The example face image displayed in Fig. \ref{fig:eximage} is taken from the LFW database \cite{lfw}. This example face image is used as the input to the pre-trained VGGFace model and features are computed before and after layer35. Fig. \ref{fig:exinput} shows the input signal for last layer (i.e. layer35). The output signal of ReLU at layer35 is displayed in Fig. \ref{fig:exoutput-relu}. In Fig. \ref{fig:exoutput-abrelu}, \ref{fig:exoutput-abrelu2}, and \ref{fig:exoutput-abrelu5}, the output signals of AB-ReLU for $\alpha$=1, 2, and 5 respectively are illustrated. For this example, $A_v^n<0$ at layer 35, it can be also observed from the Fig. \ref{fig:example} that the AB-ReLU function passes more signal as compared to the ReLU function.

\section{Experimental Setup}
In this paper, the image retrieval framework is adapted for the experiments. The face retrieval is done using the proposed AB-ReLU based VGGFace descriptor. In face retrieval, the top matching faces are returned from a database for a given query face based on the description of the faces. The best matching faces are decided based on the similarity scores between the query face and the database faces. In this work, the similarity scores are considered as the distance between the descriptor of the query face and the descriptors of the database faces. The lower distance between two feature descriptors represents more similarity amongst the corresponding face images and vice versa.   

\subsection{Distances Measures}
In image retrieval, the performance also depends upon the distance measures used for finding the similarity scores. In order to compute the performance, top few faces are retrieved. The Chi-square (Chisq) distance is used in most of the experiments in this work. The Euclidean, Cosine, Earth Mover Distance (Emd), L1, and D1 distances are also adapted to find out the more suitable distance in the current scenario \cite{ltrp}, \cite{mdlbp}.

\subsection{Evaluation Criteria}
In order to present the result of face retrieval and comparison, the standard evaluation metrics are used in this paper such as precision, recall, f-score, and retrieval rank. All the images from a database are treated as the query image (i.e., the probe sample) one by one and the rest of the images as gallery to report the average performance over full database. The average retrieval precision (ARP) and average retrieval rate (ARR) over full database are computed as the average of mean precisions (MP) and mean recalls (MR) respectively, over all categories. The MP and MR for a category is calculated as the mean of precisions and recalls respectively, by turning all the images in that category as the query one by one. The precision ($Pr$) and recall ($Re$) for a query image is calculated as follows,
\begin{equation}
\begin{aligned}
& Pr = \frac{\#Correct\_Retrieved\_Images}{\#Retrieved\_Images}\\
& Re = \frac{\#Correct\_Retrieved\_Images}{\#Similar\_Images\_In\_Database}
\end{aligned}
\end{equation}
The F-score is calculated from the ARP and ARR values with the help following equation,
\begin{equation}
F-score=2 \times \frac{ARP\times ARR}{ARP+ARR}
\end{equation}
In order to test the accurate rank of correctly retrieved faces, the average normalized modified retrieval rank (ANMRR) metric is adapted \cite{lbdp}. The better retrieval performance is inferred from the higher values of ARP, ARR and F-Score, and lower value of the ANMRR and vice-versa.

\subsection{Databases Used}
Seven challenging, unconstrained and robust face databases are used to demonstrate the effectiveness of the proposed AB-ReLU based VGGFace descriptor: PaSC \cite{pasc}, LFW \cite{lfw}, PubFig \cite{pubfig}, FERET \cite{feret}, \cite{feret1}, AR \cite{ar,ar1}, ExYaleB \cite{yale}, \cite{yale1} and PolyUNIR \cite{polyunir}. Viola Jones object detection method \cite{viola} is adopted in this paper to detect and crop the face regions in the images. The faces are resized to $224\times224$ and `zerocenter' normalization is applied before feeding to the proposed AB-ReLU based VGGFace model. 

The PaSC still image face database consists of 9376 images from 293 individuals with 32 images per individual \cite{pasc}. The PaSC database has the variations like blur, pose, and illumination and regarded as one of the difficult databases. This database has 8718 faces after face detection using Viola Jones detector. In current scenario, the unconstrained face retrieval is very demanding due to the increasing number of faces over the Internet. In this paper, LFW \cite{lfw} and PubFig \cite{pubfig} databases are considered. These two databases have the images from the Internet with several unconstrained variations, such as pose, lighting, expression, scene, camera, etc. In the image retrieval framework, it is required to retrieve more than one (typically 5, 10, etc.) top matching images. To achieve this sufficient number of images should be available for each category in the database. Considering this fact, individuals having at least 20 images are taken in the LFW database (i.e. 2984 faces from 62 individuals) \cite{lfw}. The Public Figure database (i.e., PubFig) consists 6472 faces from 60 individuals \cite{pubfig}. Following the URLs given in the PubFig face database, the images are downloaded directly from the Internet after removing the dead URLs. 

In order to experiment with the robustness of the descriptor, FERET, AR and Extended Yale B face databases are used. ``Portions of the research in this paper use the FERET database of facial images collected under the FERET program, sponsored by the DOD Counterdrug Technology Development Program Office" \cite{feret,feret1}. The cropped version of the Color-FERET database having 4053 faces from 141 people (only subjects having at least 20 faces) is considered in this work. Several variations like the expression and pose (13 different poses) are present in the FERET database. The cropped version of the AR face database is also used for the experiments \cite{ar,ar1}. The AR database has the masking effect where some portions of the face are occluded along with the illumination and color effect. Total 2600 face images are available for 100 people in AR database. Extended Yale B (ExYaleB) database is created with severe amount of illumination differences (i.e. 64 types of illuminations) \cite{yale,yale1}. Total 2432 cropped faces from 38 persons with 64 faces per person are present in the ExYaleB database for the face retrieval. The description of PolyUNIR dataset is provided section 5.4.

\section{Results and Discussions}

\begin{table*}[!t]
\caption{Average Retrieval Precision, ARP(\%) for topmost match using AB-ReLU and VGGFace based descriptors over the PaSC, LFW, PubFig, FERET, AR and ExYaleB databases. It is also equivalent to the rank-1 accuracy. The results for best performing descriptor for a database is highlighted in bold. Here, 35R, 33R and 30R represent the ReLU features at layer no. 35, 33 and 30 respectively; 35AR, 33AR and 30AR represent the AB-ReLU features for average biased parameter $\alpha=1$ at layer no. 35, 33 and 30 respectively; 35AR2 and 33AR2 represent the AB-ReLU features for average biased parameter $\alpha=2$ at layer no. 35 and 33 respectively; 35AR5 and 33AR5 represent the AB-ReLU features for average biased parameter $\alpha=5$ at layer no. 35 and 33 respectively; 33AR\_35 represents the ReLU features at layer 35 with the AB-ReLU operator at layer 33; and 33,35AR represents the AB-ReLU features at layer 35 with the AB-ReLU operator at layer 33.}
\label{arp1}
\begin{center}
\begin{tabular}{|m{.95cm}|m{.8cm}|m{.8cm}|m{.8cm}|m{.8cm}|m{.8cm}|m{.8cm}|m{.8cm}|m{.8cm}|m{.95cm}|m{.95cm}|m{.8cm}|m{.8cm}|m{.95cm}|}
\hline
Database & 35R & 35AR & 35AR2 & 35AR5 & 33R & 33AR & 33AR2 & 33AR5 & 33AR\_35 & 33,35AR & 30R & 30AR & 30AR\_35 \\
\hline\hline
PaSC & 93.06 & 93.88 & \textbf{93.89} & 93.83 & 93.36 & 93.82 & 93.88 & 93.84 & 92.96 & 93.04 & 91.95 & 86.98 & 93.02 \\
LFW & 99.10 & \textbf{99.53} & 99.31 & 99.24 & 99.21 & 99.36 & 99.32 & 99.37 & 99.22 & 99.14 & 98.05 & 94.82 & 99.30 \\
PubFig & 98.22 & 98.35 & \textbf{98.59} & 98.54 & 98.25 & 98.32 & 98.52 & 98.43 & 98.08 & 98.09 & 94.55 & 91.76 & 97.63 \\
FERET & 95.64 & \textbf{95.87} & 95.56 & 95.42 & 94.79 & 94.35 & 94.22 & 93.57 & 95.74 & 95.74 & 91.89 & 92.94 & 95.70 \\
AR & 99.73 & 99.77 & 99.81 & 99.81 & 99.85 & 99.81 & 99.81 & 99.81 & 99.77 & 99.77 & \textbf{99.96} & \textbf{99.96} & 99.77 \\
ExYaleB & 85.77 & 86.39 & 86.27 & 85.90 & 86.92 & 86.55 & 86.18 & 85.53 & 85.90 & 85.81 & \textbf{92.52} & \textbf{92.52} & 86.72 \\
\hline
\end{tabular}
\end{center}
\end{table*}

\begin{table*}[!t]
\caption{ARP(\%) for 5 numbers of retrieved images using AB-ReLU and VGGFace based descriptors over the PaSC, LFW, PubFig, FERET, AR and ExYaleB databases. The results for the best descriptor for a database is highlighted in bold.}
\label{arp5}
\begin{center}
\begin{tabular}{|m{.95cm}|m{.8cm}|m{.8cm}|m{.8cm}|m{.8cm}|m{.8cm}|m{.8cm}|m{.8cm}|m{.8cm}|m{.95cm}|m{.95cm}|m{.8cm}|m{.8cm}|m{.95cm}|}
\hline
Database & 35R & 35AR & 35AR2 & 35AR5 & 33R & 33AR & 33AR2 & 33AR5 & 33AR\_35 & 33,35AR & 30R & 30AR & 30AR\_35 \\
\hline\hline
PaSC & 87.91 & 89.33 & \textbf{89.60} & 89.37 & 87.83 & 88.79 & 89.07 & 89.27 & 87.79 & 87.80 & 79.84 & 68.46 & 86.87 \\
LFW & 98.33 & 98.49 & 98.44 & 98.39 & 98.39 & 98.50 & \textbf{98.51} & \textbf{98.51} & 98.21 & 98.17 & 94.05 & 86.31 & 98.11 \\
PubFig & 96.51 & 97.04 & \textbf{97.37} & 97.30 & 96.84 & 97.19 & 97.19 & 97.13 & 96.53 & 96.53 & 90.28 & 84.19 & 95.84 \\
FERET & 88.04 & \textbf{88.46} & 88.01 & 87.94 & 84.65 & 84.73 & 84.85 & 84.75 & 87.98 & 87.98 & 65.84 & 66.75 & 86.91 \\
AR & 94.85 & 95.15 & 95.21 & 95.12 & 95.35 & 95.56 & \textbf{95.65} & 95.63 & 94.61 & 94.57 & 94.24 & 90.83 & 94.78 \\
ExYaleB & 77.31 & 77.97 & 77.71 & 77.29 & 76.28 & 76.43 & 76.17 & 75.84 & 76.97 & 76.98 & \textbf{82.71} & 81.60 & 77.58 \\
\hline
\end{tabular}
\end{center}
\end{table*}

\begin{table*}[!t]
\caption{ARP(\%) for 10 numbers of retrieved images using AB-ReLU and VGGFace based descriptors over the PaSC, LFW, PubFig, FERET, AR and ExYaleB databases. The results for the best descriptor for a database is highlighted in bold.}
\label{arp}
\begin{center}
\begin{tabular}{|m{.95cm}|m{.8cm}|m{.8cm}|m{.8cm}|m{.8cm}|m{.8cm}|m{.8cm}|m{.8cm}|m{.8cm}|m{.95cm}|m{.95cm}|m{.8cm}|m{.8cm}|m{.95cm}|}
\hline
Database & 35R & 35AR & 35AR2 & 35AR5 & 33R & 33AR & 33AR2 & 33AR5 & 33AR\_35 & 33,35AR & 30R & 30AR & 30AR\_35 \\
\hline\hline
PaSC & 83.11 & 85.08 & \textbf{85.39} & 85.13 & 82.79 & 83.94 & 84.37 & 84.10 & 82.92 & 82.89 & 68.51 & 54.21 & 81.48 \\
LFW & 97.34 & \textbf{97.69} & 97.52 & 97.34 & 97.45 & 97.54 & 97.63 & 97.34 & 97.17 & 97.12 & 89.14 & 77.52 & 96.87 \\
PubFig & 95.06 & 95.71 & \textbf{95.90} & 95.83 & 95.41 & 95.75 & 95.72 & 95.54 & 94.91 & 94.92 & 85.85 & 77.28 & 94.11 \\
FERET & 80.28 & \textbf{81.22} & 80.92 & 80.64 & 75.90 & 75.83 & 76.18 & 75.77 & 80.14 & 80.16 & 45.61 & 45.26 & 77.80 \\
AR & 80.93 & 81.95 & 82.05 & 82.07 & 80.47 & 81.63 & 81.95 & \textbf{82.31} & 80.47 & 80.45 & 80.76 & 73.83 & 79.29 \\
ExYaleB & 70.64 & \textbf{71.54} & 71.43 & 71.05 & 68.05 & 68.17 & 68.40 & 68.68 & 70.27 & 70.26 & 70.39 & 68.51 & 70.49 \\
\hline
\end{tabular}
\end{center}
\end{table*}

\begin{table*}[!t]
\caption{Average Retrieval Rate, ARR(\%) for 10 numbers of retrieved images using AB-ReLU and VGGFace based descriptors over the PaSC, LFW, PubFig, FERET, AR and ExYaleB databases. The results for the best descriptor for a database is highlighted in bold.}
\label{arr}
\begin{center}
\begin{tabular}{|m{.95cm}|m{.8cm}|m{.8cm}|m{.8cm}|m{.8cm}|m{.8cm}|m{.8cm}|m{.8cm}|m{.8cm}|m{.95cm}|m{.95cm}|m{.8cm}|m{.8cm}|m{.95cm}|}
\hline
Database & 35R & 35AR & 35AR2 & 35AR5 & 33R & 33AR & 33AR2 & 33AR5 & 33AR\_35 & 33,35AR & 30R & 30AR & 30AR\_35 \\
\hline\hline
PaSC & 28.06 & 28.74 & \textbf{28.83} & 28.74 & 27.95 & 28.33 & 28.48 & 28.38 & 27.99 & 27.98 & 23.09 & 18.29 & 27.50 \\
LFW & 31.48 & \textbf{31.61} & 31.53 & 31.46 & 31.51 & 31.53 & 31.57 & 31.44 & 31.42 & 31.39 & 28.31 & 24.04 & 31.31 \\
PubFig & 17.44 & 17.57 & \textbf{17.64} & 17.62 & 17.53 & 17.61 & 17.60 & 17.58 & 17.41 & 17.42 & 15.21 & 13.05 & 17.21 \\
FERET & 30.32 & \textbf{30.67} & 30.54 & 30.43 & 28.63 & 28.57 & 28.73 & 28.59 & 30.27 & 30.27 & 17.10 & 17.10 & 29.32 \\
AR & 31.13 & 31.52 & 31.56 & 31.56 & 30.95 & 31.40 & 31.52 & \textbf{31.66} & 30.95 & 30.94 & 31.06 & 28.40 & 30.50 \\
ExYaleB & 11.04 & \textbf{11.18} & 11.16 & 11.10 & 10.63 & 10.65 & 10.69 & 10.73 & 10.98 & 10.98 & 11.00 & 10.70 & 11.01 \\
\hline
\end{tabular}
\end{center}
\end{table*}

\begin{table*}[!t]
\caption{F-Score(\%) for 10 numbers of retrieved images using AB-ReLU and VGGFace based descriptors over the PaSC, LFW, PubFig, FERET, AR and ExYaleB databases. The results for the best descriptor for a database is highlighted in bold.}
\label{fscore}
\begin{center}
\begin{tabular}{|m{.95cm}|m{.8cm}|m{.8cm}|m{.8cm}|m{.8cm}|m{.8cm}|m{.8cm}|m{.8cm}|m{.8cm}|m{.95cm}|m{.95cm}|m{.8cm}|m{.8cm}|m{.95cm}|}
\hline
Database & 35R & 35AR & 35AR2 & 35AR5 & 33R & 33AR & 33AR2 & 33AR5 & 33AR\_35 & 33,35AR & 30R & 30AR & 30AR\_35 \\
\hline\hline
PaSC & 41.89 & 42.89 & \textbf{43.04} & 42.90 & 41.72 & 42.30 & 42.52 & 42.37 & 41.79 & 41.77 & 34.50 & 27.31 & 41.06 \\
LFW & 46.05 & \textbf{46.23} & 46.13 & 46.02 & 46.10 & 46.13 & 46.18 & 46.00 & 45.96 & 45.92 & 41.55 & 35.42 & 45.80 \\
PubFig & 26.86 & 27.06 & \textbf{27.17} & 27.14 & 26.98 & 27.11 & 27.09 & 27.06 & 26.82 & 26.83 & 23.57 & 20.52 & 26.52 \\
FERET & 43.46 & \textbf{43.97} & 43.79 & 43.63 & 41.05 & 40.97 & 41.20 & 40.99 & 43.38 & 43.40 & 24.55 & 24.50 & 42.05 \\
AR & 44.96 & 45.53 & 45.58 & 45.59 & 44.71 & 45.35 & 45.53 & \textbf{45.73} & 44.70 & 44.69 & 44.87 & 41.02 & 44.05 \\
ExYaleB & 19.09 & \textbf{19.33} & 19.30 & 19.20 & 18.39 & 18.42 & 18.49 & 18.56 & 18.99 & 18.99 & 19.03 & 18.52 & 19.05 \\
\hline
\end{tabular}
\end{center}
\end{table*}

\begin{table*}[!t]
\caption{Average Normalized Modified Retrieval Rank (ANMRR) in \% for 10 numbers of retrieved images using AB-ReLU and VGGFace based descriptors over the PaSC, LFW, PubFig, FERET, AR and ExYaleB databases. The results for the best performing descriptor (i.e. least ANMRR value) for is highlighted in bold.}
\label{anmrr}
\begin{center}
\begin{tabular}{|m{.95cm}|m{.8cm}|m{.8cm}|m{.8cm}|m{.8cm}|m{.8cm}|m{.8cm}|m{.8cm}|m{.8cm}|m{.95cm}|m{.95cm}|m{.8cm}|m{.8cm}|m{.95cm}|}
\hline
Database & 35R & 35AR & 35AR2 & 35AR5 & 33R & 33AR & 33AR2 & 33AR5 & 33AR\_35 & 33,35AR & 30R & 30AR & 30AR\_35 \\
\hline\hline
PaSC & 4.40 & \textbf{3.40} & 3.43 & 3.61 & 4.43 & 3.98 & 3.83 & 4.23 & 4.51 & 4.54 & 17.60 & 33.97 & 5.28 \\
LFW & 0.42 & 0.38 & 0.44 & 0.49 & 0.41 & 0.41 & 0.40 & 0.55 & 0.46 & 0.49 & 5.29 & 16.33 & \textbf{0.32} \\
PubFig & 0.85 & 0.66 & \textbf{0.58} & 0.60 & 0.72 & \textbf{0.58} & 0.62 & 0.64 & 0.91 & 0.89 & 6.65 & 14.44 & 1.25 \\
FERET & 13.01 & \textbf{12.15} & 12.50 & 12.80 & 17.24 & 17.43 & 17.15 & 17.69 & 13.12 & 13.11 & 49.50 & 49.86 & 15.31 \\
AR & 4.13 & 3.40 & 3.42 & 3.41 & 4.25 & 3.53 & 3.32 & \textbf{3.30} & 4.51 & 4.52 & 3.80 & 10.54 & 5.24 \\
ExYaleB & 10.82 & 9.98 & \textbf{9.88} & 10.22 & 13.68 & 13.59 & 13.36 & 13.12 & 11.18 & 11.17 & 10.76 & 12.93 & 10.96 \\
\hline
\end{tabular}
\end{center}
\end{table*}

\begin{table*}[!t]
\caption{ARP in \% for 10 numbers of retrieved images using VGGFace35ReLU (35R) and VGGFace35AB-ReLU (35AR) descriptors with different distance measures. Note that ARP value highlighted in bold represents the best distance over a database for AB-ReLU.}
\label{dis}
\begin{center}
\begin{tabular}{|m{1.2cm}|m{1.1cm}|m{1.3cm}|m{1.1cm}|m{1.3cm}|m{1.1cm}|m{1.3cm}|m{1.1cm}|m{1.3cm}|m{1.1cm}|m{1.3cm}|}
\hline
\multirow{2}{*}{Database} & \multicolumn{2}{|m{1.3cm}|}{Euclidean} & \multicolumn{2}{|m{1.3cm}|}{Cosine} & \multicolumn{2}{|m{1.3cm}|}{L1} & \multicolumn{2}{|m{1.3cm}|}{D1} & \multicolumn{2}{|m{1.3cm}|}{Chi-square} \\
\cline{2-11}
& ReLU & AB-ReLU & ReLU & AB-ReLU & ReLU & AB-ReLU & ReLU & AB-ReLU & ReLU & AB-ReLU \\
\hline\hline
\vspace{1mm}
PaSC & 79.38 & 84.89 & 82.83 & 84.84 & 82.96 & 85.01 & 82.96 & 85.01 & 83.11 & \textbf{85.08} \\
LFW & 95.06 & 97.65 & 97.21 & 97.64 & 97.30 & 97.66 & 97.30 & 97.66 & 97.33 & \textbf{97.69} \\
PubFig & 92.56 & 95.64 & 94.90 & 95.63 & 94.99 & 95.67 & 95.00 & 95.67 & 95.06& \textbf{95.71} \\
FERET & 78.72 & \textbf{81.32} & 79.88 & 81.18 & 80.21 & 81.25 & 80.23 & 81.25 & 80.28 & 81.22 \\
AR & 79.25 & 81.80 & 80.51 & 81.78 & 80.85 & 81.93 & 80.85 & 81.93 & 80.93 & \textbf{81.95} \\
ExYaleB & 69.98 & 71.45 & 70.27 & 71.34 & 70.53 & 71.48 & 70.51 & 71.48 & 70.64 & \textbf{71.54} \\
\hline
\end{tabular}
\end{center}
\end{table*}

In this work, the content based image retrieval framework is adopted for the experiments and comparison. In this section, first result comparison is presented by fixing the similarity measure as Chi-square distance, then the performance of proposed VGGFace35AB-ReLU descriptor is tested with different similarity measures, then the training is performed by replacing all ReLU with AB-ReLU and validation error is compared over six datasets, after that the experiment over a large scale dataset is performed, and finally the results are compared with other functions.

\subsection{Results Comparison}
Several VGGFace descriptors with AB-ReLU at different layers such as VGGFace35ReLU (35R), VGGFace35AB-ReLU (35AR), VGGFace35AB-ReLU2 (35AR2), VGGFace35AB-ReLU5 (35AR5), VGGFace33ReLU (33R), VGGFace33AB-ReLU (33AR), VGGFace33AB-ReLU2 (33AR)2, VGGFace33AB-ReLU5 (33AR5), VGGFace33AB-ReLU\_35 (33AR\_35), VGGFace33,35AB-ReLU (33,35AR), VGGFace30AB-ReLU (30AR), VGGFace30AB-ReLU (30AR), and VGGFace30AB-ReLU\_35 (30AR\_35), etc. are used for the experiments. Here, 35R, 33R and 30R represent the ReLU features at layer no. 35, 33 and 30 respectively; 35AR, 33AR and 30AR represent the AB-ReLU features for average biased parameter $\alpha=1$ at layer no. 35, 33 and 30 respectively; 35AR2 and 33AR2 represent the AB-ReLU features for average biased parameter $\alpha=2$ at layer no. 35 and 33 respectively; 35AR5 and 33AR5 represent the AB-ReLU features for average biased parameter $\alpha=5$ at layer no. 35 and 33 respectively; 30AR\_35 represents the ReLU features at layer 35 with the AB-ReLU operator at layer 30; and 33,35AR represents the AB-ReLU features at layer 35 with the AB-ReLU operator at layer 33.

The average retrieval precision (ARP) for the topmost match (i.e. Rank-1 Accuracy) is illustrated in Table \ref{arp1} over the PaSC, LFW, PubFig, FERET, AR, and ExYaleB databases. It is observed from Table \ref{arp1} that the performance of 35AR and 35AR2 is better, mainly over the unconstrained databases, whereas the performance of 30R and 30AR is better over robust databases like AR and ExYaleB. It is also noted that the performance of AB-ReLU (35AR) is improved as compared to the ReLU (35R) at layer 35. The AB-ReLU is better than ReLU over unconstrained databases like PaSC, LFW and PubFig at layer 33 (see the comparison between 33R and 33AR in Table \ref{arp1}). Interestingly, the AB-ReLU performs better or similar to the ReLU over robust databases such as FERET, AR and ExYaleB at layer 30 (Please see the comparative results of 30R and 30AR in Table \ref{arp1}). The reason behind such a performance is due to more abstract level feature learning towards the last layer. So, changing the learned features at early layers affects feature extraction adversely. 

The FaceNet approach \cite{facenet}  has reported the best accuracy of 99.63\% over LFW database. Whereas, the proposed method has achieved the best accuracy of 99.53\% as shown in Table II, which is very close to the FaceNet result. However, the FaceNet approach has used 22-layer architecture and the proposed approach uses the 16-layer pre-trained VGGFace architecture. One of the major differences with FaceNet approach is that it uses triplet loss instead of softmax loss are used in the proposed method. 

Table \ref{arp5} listed the ARP values when 5 best faces are retrieved. This result shows that the performance is generally better for the parameter $\alpha=2$, i.e. 35AR2 and 33AR2. The picture is clear from Table \ref{arp}, where ARP is reported for 10 retrieved images. Descriptors constructed using AB-ReLU at the last layer (i.e. layer35) are superior for most of the databases. However, optimum result is achieved at $33^{rd}$ layer over AR database. One possible reason is that the trained faces of VGGFace database are not masked. The result in Table \ref{arp} confirms that AB-ReLU is better suited for the descriptor as compared to ReLU at layer35. The ARR and F-Score values are summarized in Table \ref{arr} and Table \ref{fscore} respectively, for 10 numbers of retrieved images. The similar trend is observed in the results of ARR and F-Score that 35AR and 35AR2 are the best performing VGGFace network based descriptors. Some variations can be seen in the ANMRR results for same 10 best matching retrieved images in Table \ref{anmrr} as compared to the ARP, ARR and F-Score because ANMRR penalizes the rank heavily for false positive retrieved images. Still 35AR is better over PaSC and FERET databases and 35AR2 is better over PubFig and ExYaleB databases. It can be noticed that the F-Score and ANMRR over LFW database is best for 35AR and 30AR\_35 descriptors respectively. It means that, the true positive rate for 30AR\_35 descriptor over LFW database is low as compared to 35AR descriptor, whereas the retrieved faces using 30AR\_35 are closer to the query face in terms of their ranks.

\begin{figure*}
\centering
  \begin{subfigure}{.48\textwidth}
    \centering
    \includegraphics[width=1\linewidth]{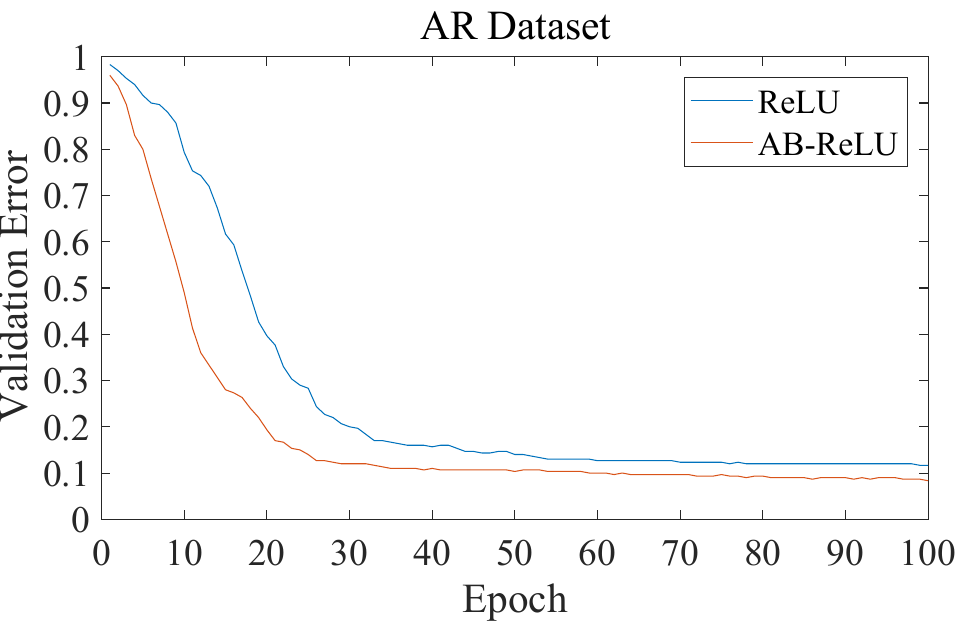}
    \caption{Validation error over AR database}
    \label{fig:val-err-ar}
  \end{subfigure}%
  \begin{subfigure}{.48\textwidth}
    \centering
    \includegraphics[width=1\linewidth]{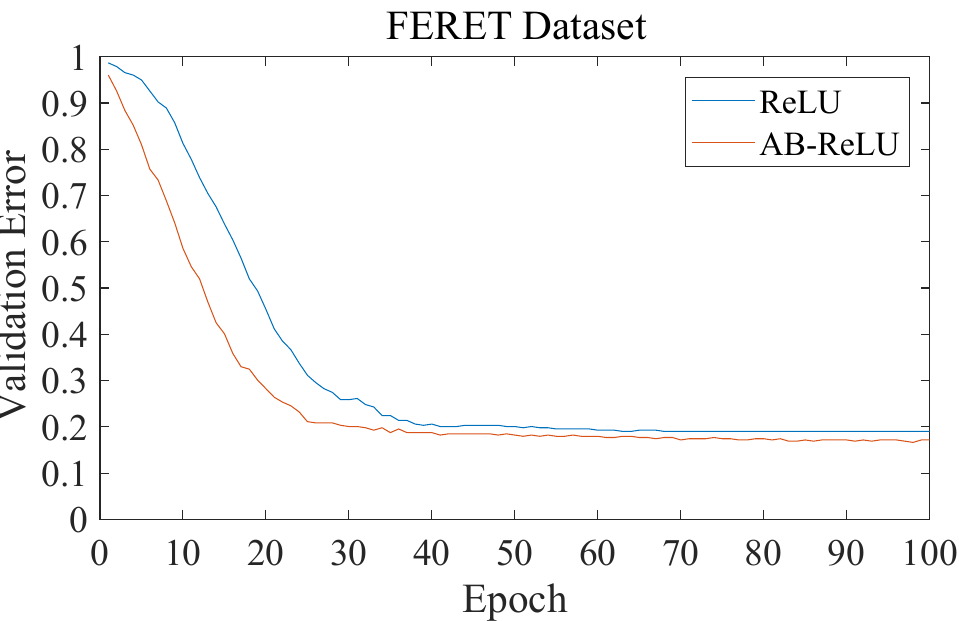}
    \caption{Validation error over FERET database}
    \label{fig:val-err-feret}
  \end{subfigure}
  \begin{subfigure}{.48\textwidth}
    \centering
    \includegraphics[width=1\linewidth]{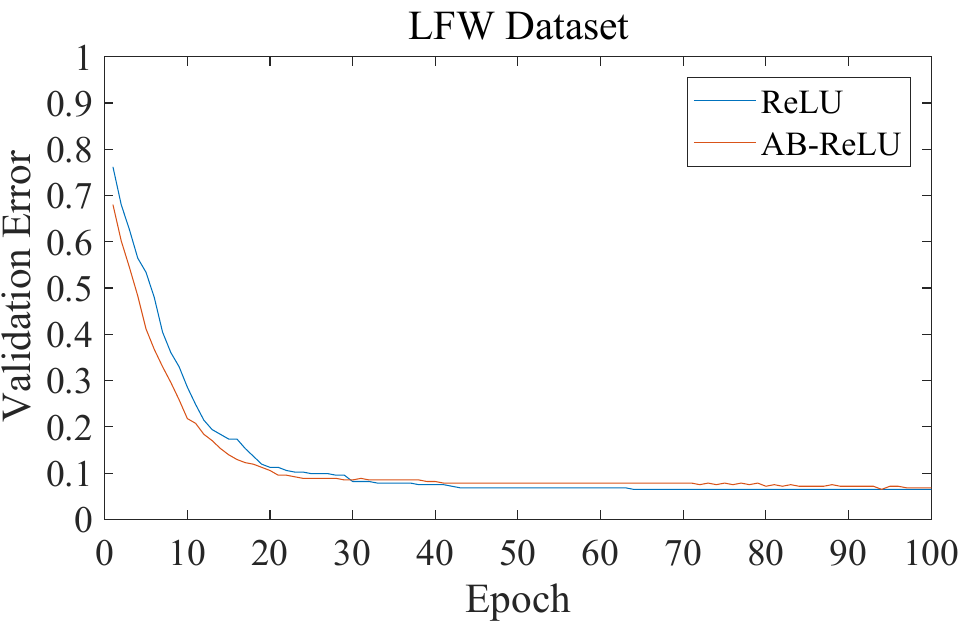}
    \caption{Validation error over LFW database}
    \label{fig:val-err-lfw}
  \end{subfigure}%
  \begin{subfigure}{.48\textwidth}
    \centering
    \includegraphics[width=1\linewidth]{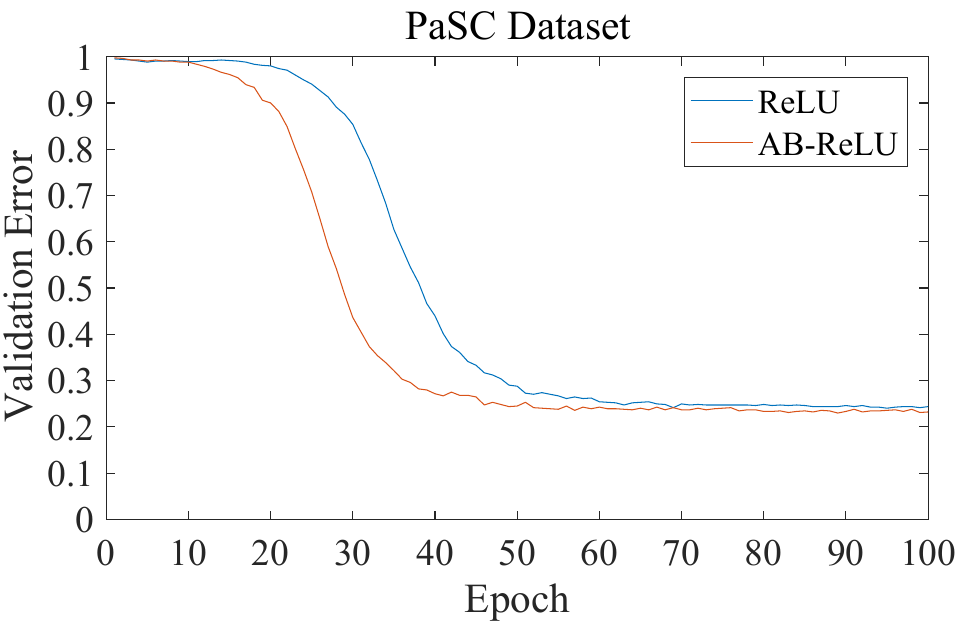}
    \caption{Validation error over PaSC database}
    \label{fig:val-err-pasc}
  \end{subfigure}
  \begin{subfigure}{.48\textwidth}
    \centering
    \includegraphics[width=1\linewidth]{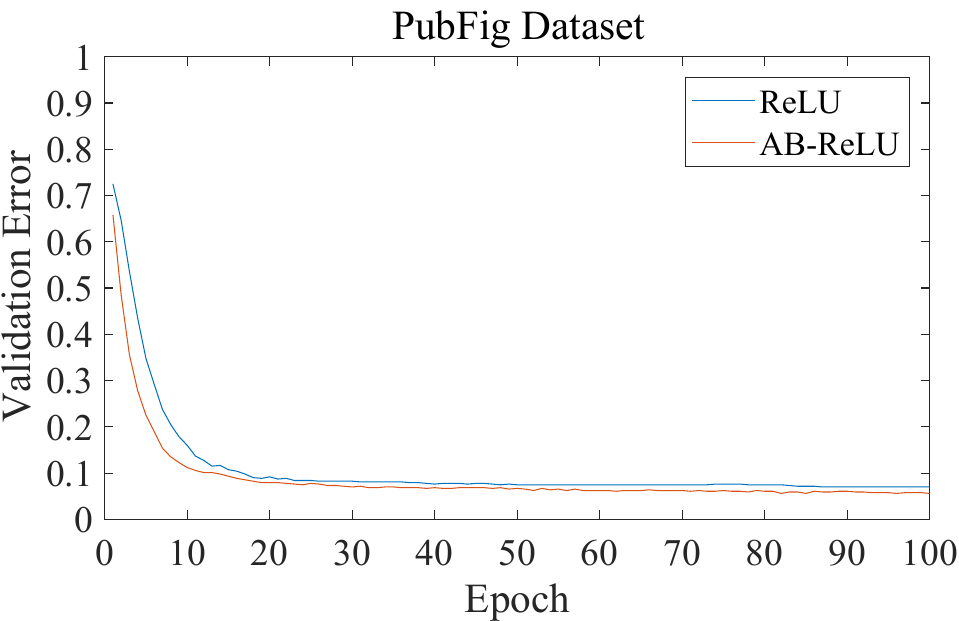}
    \caption{Validation error over PubFig database}
    \label{fig:val-err-pubfig}
  \end{subfigure}%
  \begin{subfigure}{.48\textwidth}
    \centering
    \includegraphics[width=1\linewidth]{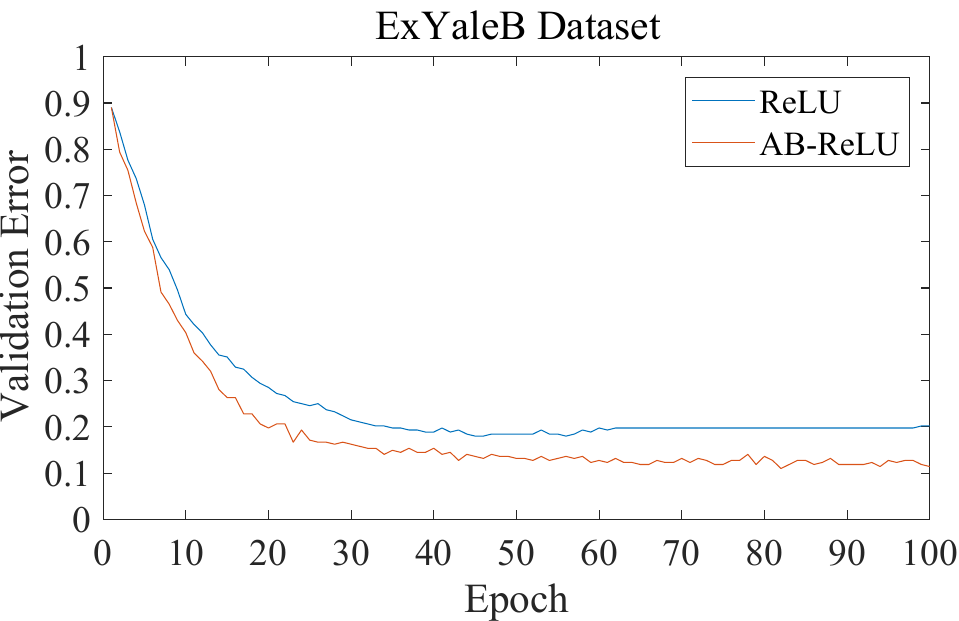}
    \caption{Validation error over ExYaleB database}
    \label{fig:val-err-yale}
  \end{subfigure}%
  \caption{Validation error over different databases with ReLU and AB-ReLU ($\alpha=0.1$) across all layers.}
  \label{fig:val-err}
\end{figure*}

\subsection{Effect of Similarity Measure}
In the comparison results of the previous subsection, Chi-square distance was adopted as the similarity measure. This experiment is conducted to reveal the best suitable similarity measure for the proposed descriptor. The ARP values using VGGFace35AB-ReLU (i.e. 35AR) descriptor over each database are presented in Table \ref{dis}. The ARP values using VGGFace35ReLU (i.e. 35R) are listed to depict the performance comparison between ReLU and AB-ReLU using different distance measure. In this experiment, 10 top matching images are retrieved with different distances. The Euclidean, Cosine, L1, D1 and Chi-square distances are experimented and reported in Table \ref{dis}. It is noticed that the Chi-square distance is better suited for each database except the FERET database. It is also observed across the Table \ref{dis} that the AB-ReLU consistently performs better than ReLU using each distance measure.

\subsection{Comparative Study Replacing all ReLU with AB-ReLU}
We also did experiments after replacing all ReLUs with AB-ReLUs in pre-trained VGGFace model. By doing so, the performance decreased as changing the values of pre-trained model in initial layers impacts the final outcome adversely. Thus, in order to generalize the feature extraction using AB-ReLU, we retained the VGGFace model using the Pre-trained weights for initialization. The training and validation are performed over each database. In this experiment, following is the hyper-parameter settings: $\alpha = 0.1$ for AB-ReLU, Weight Decay = 0.0001, Optimizer: SGDM, Momentum = 0.95, Batch Size = 16, No. of Epochs = 100, and Learning rate = 0.000001 for all epochs. The validation errors of ReLU and AB-ReLU over different databases are shown in Fig. \ref{fig:val-err}.  It is evident from Fig. \ref{fig:val-err} that the network achieves lower validation error when it has been trained with AB-ReLU. Moreover, the convergence using AB-ReLU is faster than ReLU.

\begin{figure}
\centering
    \includegraphics[width=1\columnwidth]{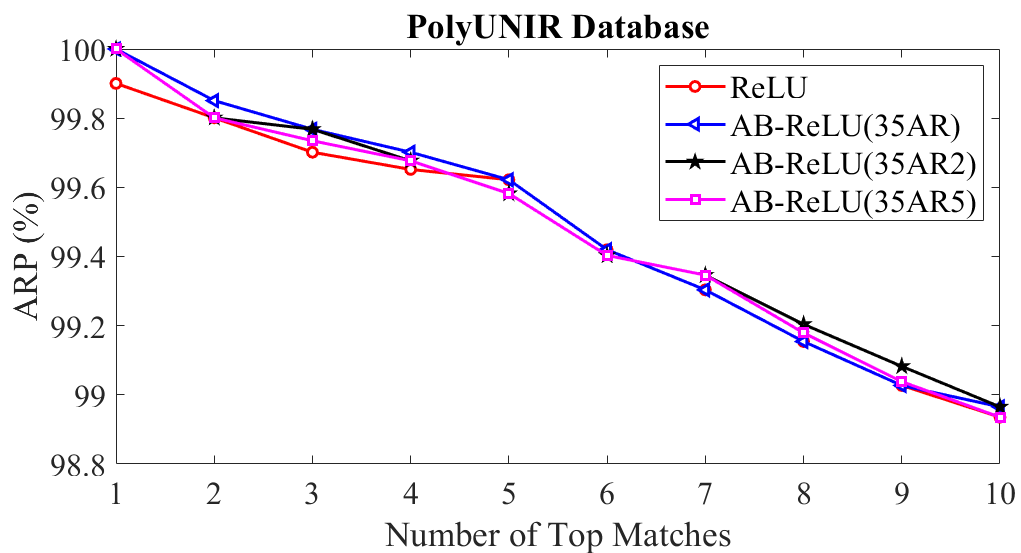}
  \caption{The ARP (\%) using the feature vectors of layer 35 generated using ReLU (i.e., 35R) and AB-ReLU (i.e., 35AR, 35AR2 and 35AR5) methods.}
  \label{fig:polyunir}
\end{figure}

\subsection{Experiment over Large Scale PolyUNIR Face Dataset}
In order to judge the performance of the proposed feature extractor on the large scale face dataset, we also performed an experiment over PolyUNIR face dataset\footnote{http://www.comp.polyu.edu.hk/~biometrics/NIRFace/polyudb\_face.htm} \cite{polyunir}. This dataset contains 38,981 face images from 335 subjects captured using a near infrared sensor. In this experiment, we have considered 1005 face images as the query image. The retrieval results over the PolyUNIR face dataset in terms of the ARP are reported in Fig. \ref{fig:polyunir}. The ReLU and three variants of AB-ReLU such as 35AR, 35AR2 and 35AR5 are used to extract the features from layer 35. The improved performance of AB-ReLU over the PolyUNIR face dataset confirms its suitability over large scale dataset.

\begin{table}[!t]
\caption{The results comparison between Sigmoid, ReLU, Leaky ReLU (LReLU with 0.01 leaky factor), Flexible ReLU (FReLU with 0.2 flexible factor) and the proposed AB-ReLU at layer 35 in terms of the ARP in \% for 10 numbers of retrieved images using Chi-square distance measure. Note that ARP value highlighted in bold represents the best result for a database.}
\label{table:lrelu}
\begin{center}
\begin{tabular}{|m{1.2cm}|m{0.9cm}|m{0.9cm}|m{0.9cm}|m{0.9cm}|m{1.3cm}|m{1.2cm}|}
\hline
Database & Sigmoid & ReLU & LReLU & FReLU & AB-ReLU \\
\hline\hline
PaSC & 29.71 & 83.11 & 83.56 & 81.65 & \textbf{85.08} \\
LFW & 53.78 & 97.34 & 97.38 & 97.07 & \textbf{97.69} \\
PubFig & 50.66 & 95.06 & 95.39 & 94.78 & \textbf{95.71} \\
FERET & 39.60 & 80.28 & 80.39 & 79.80 & \textbf{81.22} \\
AR & 32.34 & 80.93 & 81.13 & 81.21 & \textbf{81.95} \\
ExYaleB & 43.84 & 70.64 & 70.66 & 69.80 & \textbf{71.54} \\
PolyUNIR & 93.05 & 98.54 & 98.97 & 98.91 &  \textbf{98.97} \\
\hline
\end{tabular}
\end{center}
\end{table}

\subsection{Comparison with Other Activation Functions}
We have conducted this experiment to compare the results with other activation functions such as Sigmoid, ReLU \cite{relu}, Leaky ReLU (LReLU) \cite{lrelu} and non-trainable Flexible ReLU (FReLU) \cite{frelu}. The Sigmoid is a traditional activation function which squashes the input and produces the output between 0 and 1. However, it suffers with the saturation problem. The LReLU and FReLU are the variants of ReLU. The leaky factor for LReLU is considered as 0.01. The flexibility factor in FReLU is considered as 0.2. All the features using these functions are considered from the layer 35 of VGGNet. Note that the Sigmoid, ReLU, FReLU and AB-ReLU consist of the positive values only, whereas the LReLU has negative values also. Due to negative values most distance measures fail for LReLU. In order to avoid this problem, we have considered the absolute value of LReLU features. The Chi-square distance measure is considered in this comparison. The experimental setting for the PolyUNIR face dataset is same as in the previous experiment. The results in terms of the ARP values for 10 numbers of retrieved images are summarized in Table \ref{table:lrelu} using Sigmoid, ReLU, LReLU, FReLU and the proposed AB-ReLU functions. These results confirm the superiority of the proposed Average Biased ReLU over Sigmoid, ReLU, Leaky ReLU and Flexible ReLU. Due to the saturation property of Sigmoid function, it loses the discriminative ability. Thus, the results of Sigmoid function are very bad. The LReLU and FReLU do not utilize the distributive property of the generated features which is incorporated in the proposed AB-ReLU function.

\section{Conclusion}
In this paper, an average biased rectified linear unit (AB-ReLU) is proposed for the image representation using CNN model. The AB-ReLU is data dependent and adjust the threshold based on the positive and negative data dominance. It considers the average of the input volume to adjust the input volume itself. The advantage of AB-ReLU is that it allows the important negative signals as well as blocks the irrelevant positive signals based on the nature of the input volume. The AB-ReLU is applied over pre-trained VGGFace model at last few layers by replacing the conventional ReLU layers. The face retrieval experiments are conducted to test the performance of AB-ReLU based VGGFace descriptor. Seven challenging face datasets are considered, including three unconstrained datasets, three robust datasets and one large scale dataset. Based on the experimental analysis, it is concluded that AB-ReLU is better suited than the simple ReLU at the last layer for a pre-trained CNN model based feature description. Favorable performance is reported in both unconstrained as well as robust scenarios. It is also found that the Chi-square distance is better suited for the proposed descriptor for face retrieval. It is also noticed that the network can be trained by replacing all ReLU and AB-ReLU with improved validation. 
The suitability of the proposed method is also confirmed through large scale dataset. The AB-ReLU also outperforms the activation functions such as Sigmoid, ReLU, Leaky ReLU and Flexible ReLU using pre-trained VGGFace model for the face retrieval task.

\section*{Acknowledgement}
This research is funded by IIIT Sri City, India through the Faculty Seed Research Grant.
We gratefully acknowledge the support of NVIDIA Corporation with the donation of the GeForce Titan X Pascal used for this research.

{\small
\bibliographystyle{IEEEtran}
\bibliography{egbib}

\begin{thebibliography}{10}
\providecommand{\url}[1]{{#1}}
\providecommand{\urlprefix}{URL }
\expandafter\ifx\csname urlstyle\endcsname\relax
  \providecommand{\doi}[1]{DOI~\discretionary{}{}{}#1}\else
  \providecommand{\doi}{DOI~\discretionary{}{}{}\begingroup
  \urlstyle{rm}\Url}\fi

\bibitem{lbpface}
Ahonen, T., Hadid, A., Pietikainen, M.: Face description with local binary
  patterns: Application to face recognition.
\newblock IEEE transactions on pattern analysis and machine intelligence
  \textbf{28}(12), 2037--2041 (2006)

\bibitem{bansal2017s}
Bansal, A., Castillo, C., Ranjan, R., Chellappa, R.: The do's and don'ts for
  cnn-based face verification.
\newblock arXiv preprint arXiv:1705.07426  (2017)

\bibitem{pasc}
Beveridge, J.R., Phillips, P.J., Bolme, D.S., Draper, B.A., Givens, G.H., Lui,
  Y.M., Teli, M.N., Zhang, H., Scruggs, W.T., Bowyer, K.W., et~al.: The
  challenge of face recognition from digital point-and-shoot cameras.
\newblock In: Biometrics: Theory, Applications and Systems (BTAS), 2013 IEEE
  Sixth International Conference on, pp. 1--8. IEEE (2013)

\bibitem{lghp}
Chakraborty, S., Singh, S., Chakraborty, P.: Local gradient hexa pattern: A
  descriptor for face recognition and retrieval.
\newblock IEEE Transactions on Circuits and Systems for Video Technology
  (2016)

\bibitem{csqp}
Chakraborty, S., Singh, S.K., Chakraborty, P.: Centre symmetric quadruple
  pattern: A novel descriptor for facial image recognition and retrieval.
\newblock Pattern Recognition Letters  (2017)

\bibitem{ldgp}
Chakraborty, S., Singh, S.K., Chakraborty, P.: Local directional gradient
  pattern: a local descriptor for face recognition.
\newblock Multimedia Tools and Applications \textbf{76}(1), 1201--1216 (2017)

\bibitem{clevert2015rectified}
Clevert, D.A., Mayr, A., Unterthiner, T., Hochreiter, S.: Rectified factor
  networks.
\newblock In: Advances in neural information processing systems, pp. 1855--1863
  (2015)

\bibitem{elu}
Clevert, D.A., Unterthiner, T., Hochreiter, S.: Fast and accurate deep network
  learning by exponential linear units (elus).
\newblock arXiv preprint arXiv:1511.07289  (2015)

\bibitem{imagenet}
Deng, J., Dong, W., Socher, R., Li, L.J., Li, K., Fei-Fei, L.: Imagenet: A
  large-scale hierarchical image database.
\newblock In: Computer Vision and Pattern Recognition, 2009. CVPR 2009. IEEE
  Conference on, pp. 248--255. IEEE (2009)

\bibitem{fdlbp}
Dubey, S.R.: Face retrieval using frequency decoded local descriptor.
\newblock Multimedia Tools and Applications \textbf{78}(12), 16411--16431
  (2019)

\bibitem{ldrp}
Dubey, S.R.: Local directional relation pattern for unconstrained and robust
  face retrieval.
\newblock Multimedia Tools and Applications  (2019)

\bibitem{ldop}
Dubey, S.R., Mukherjee, S.: Ldop: Local directional order pattern for robust
  face retrieval.
\newblock arXiv preprint arXiv:1803.07441  (2018)

\bibitem{iold}
Dubey, S.R., Singh, S.K., Singh, R.K.: Rotation and illumination invariant
  interleaved intensity order-based local descriptor.
\newblock IEEE Transactions on Image Processing \textbf{23}(12), 5323--5333
  (2014)

\bibitem{ldep}
Dubey, S.R., Singh, S.K., Singh, R.K.: Local diagonal extrema pattern: a new
  and efficient feature descriptor for ct image retrieval.
\newblock IEEE Signal Processing Letters \textbf{22}(9), 1215--1219 (2015)

\bibitem{lwp}
Dubey, S.R., Singh, S.K., Singh, R.K.: Local wavelet pattern: A new feature
  descriptor for image retrieval in medical ct databases.
\newblock IEEE Transactions on Image Processing \textbf{24}(12), 5892--5903
  (2015)

\bibitem{lbdp}
Dubey, S.R., Singh, S.K., Singh, R.K.: Local bit-plane decoded pattern: a novel
  feature descriptor for biomedical image retrieval.
\newblock IEEE Journal of Biomedical and Health Informatics \textbf{20}(4),
  1139--1147 (2016)

\bibitem{mdlbp}
Dubey, S.R., Singh, S.K., Singh, R.K.: Multichannel decoded local binary
  patterns for content-based image retrieval.
\newblock IEEE Transactions on Image Processing \textbf{25}(9), 4018--4032
  (2016)

\bibitem{geexploiting}
Ge, Y., Jiang, S., Xu, Q., Jiang, C., Ye, F.: Exploiting representations from
  pre-trained convolutional neural networks for high-resolution remote sensing
  image retrieval.
\newblock Multimedia Tools and Applications pp. 1--27 (2018)

\bibitem{yale}
Georghiades, A.S., Belhumeur, P.N., Kriegman, D.J.: From few to many:
  Illumination cone models for face recognition under variable lighting and
  pose.
\newblock IEEE transactions on pattern analysis and machine intelligence
  \textbf{23}(6), 643--660 (2001)

\bibitem{he2015delving}
He, K., Zhang, X., Ren, S., Sun, J.: Delving deep into rectifiers: Surpassing
  human-level performance on imagenet classification.
\newblock In: Proceedings of the IEEE international conference on computer
  vision, pp. 1026--1034 (2015)

\bibitem{resnet}
He, K., Zhang, X., Ren, S., Sun, J.: Deep residual learning for image
  recognition.
\newblock In: Proceedings of the IEEE conference on computer vision and pattern
  recognition, pp. 770--778 (2016)

\bibitem{lfw}
Huang, G.B., Ramesh, M., Berg, T., Learned-Miller, E.: Labeled faces in the
  wild: A database for studying face recognition in unconstrained environments.
\newblock Tech. rep., Technical Report 07-49, University of Massachusetts,
  Amherst (2007)

\bibitem{alexnet}
Krizhevsky, A., Sutskever, I., Hinton, G.E.: Imagenet classification with deep
  convolutional neural networks.
\newblock In: Advances in neural information processing systems, pp. 1097--1105
  (2012)

\bibitem{pubfig}
Kumar, N., Berg, A.C., Belhumeur, P.N., Nayar, S.K.: Attribute and simile
  classifiers for face verification.
\newblock In: Computer Vision, 2009 IEEE 12th International Conference on, pp.
  365--372. IEEE (2009)

\bibitem{yale1}
Lee, K.C., Ho, J., Kriegman, D.J.: Acquiring linear subspaces for face
  recognition under variable lighting.
\newblock IEEE Transactions on pattern analysis and machine intelligence
  \textbf{27}(5), 684--698 (2005)

\bibitem{li2019piecewise}
Li, Y., Wan, L., Fu, T., Hu, W.: Piecewise supervised deep hashing for image
  retrieval.
\newblock Multimedia Tools and Applications pp. 1--21 (2019)

\bibitem{liu2017fusion}
Liu, P., Guo, J.M., Wu, C.Y., Cai, D.: Fusion of deep learning and compressed
  domain features for content-based image retrieval.
\newblock IEEE Transactions on Image Processing \textbf{26}(12), 5706--5717
  (2017)

\bibitem{ma2019multimedia}
Ma, X., Jiang, X.: Multimedia image quality assessment based on deep feature
  extraction.
\newblock Multimedia Tools and Applications pp. 1--12 (2019)

\bibitem{lrelu}
Maas, A.L., Hannun, A.Y., Ng, A.Y.: Rectifier nonlinearities improve neural
  network acoustic models.
\newblock In: Proc. ICML, vol.~30 (2013)

\bibitem{ar}
Martinez, A.M.: The ar face database.
\newblock CVC technical report  (1998)

\bibitem{ar1}
Mart{\'\i}nez, A.M., Kak, A.C.: Pca versus lda.
\newblock IEEE transactions on pattern analysis and machine intelligence
  \textbf{23}(2), 228--233 (2001)

\bibitem{ltrp}
Murala, S., Maheshwari, R., Balasubramanian, R.: Local tetra patterns: a new
  feature descriptor for content-based image retrieval.
\newblock IEEE Transactions on Image Processing \textbf{21}(5), 2874--2886
  (2012)

\bibitem{relu}
Nair, V., Hinton, G.E.: Rectified linear units improve restricted boltzmann
  machines.
\newblock In: Proceedings of the 27th international conference on machine
  learning (ICML-10), pp. 807--814 (2010)

\bibitem{vggface}
Parkhi, O.M., Vedaldi, A., Zisserman, A., et~al.: Deep face recognition.
\newblock In: BMVC, vol.~1, p.~6 (2015)

\bibitem{feret1}
Phillips, P.J., Moon, H., Rizvi, S.A., Rauss, P.J.: The feret evaluation
  methodology for face-recognition algorithms.
\newblock IEEE Transactions on pattern analysis and machine intelligence
  \textbf{22}(10), 1090--1104 (2000)

\bibitem{feret}
Phillips, P.J., Wechsler, H., Huang, J., Rauss, P.J.: The feret database and
  evaluation procedure for face-recognition algorithms.
\newblock Image and vision computing \textbf{16}(5), 295--306 (1998)

\bibitem{frelu}
Qiu, S., Xu, X., Cai, B.: Frelu: Flexible rectified linear units for improving
  convolutional neural networks.
\newblock In: 2018 24th International Conference on Pattern Recognition (ICPR),
  pp. 1223--1228. IEEE (2018)

\bibitem{facenet}
Schroff, F., Kalenichenko, D., Philbin, J.: Facenet: A unified embedding for
  face recognition and clustering.
\newblock In: Proceedings of the IEEE Conference on Computer Vision and Pattern
  Recognition, pp. 815--823 (2015)

\bibitem{shamsolmoali2019high}
Shamsolmoali, P., Jain, D.K., Zareapoor, M., Yang, J., Alam, M.A.:
  High-dimensional multimedia classification using deep cnn and extended
  residual units.
\newblock Multimedia Tools and Applications \textbf{78}(17), 23867--23882
  (2019)

\bibitem{sharma2015identity}
Sharma, S., Dubey, S.R., Singh, S.K., Saxena, R., Singh, R.K.: Identity
  verification using shape and geometry of human hands.
\newblock Expert Systems with Applications \textbf{42}(2), 821--832 (2015)

\bibitem{vggnet}
Simonyan, K., Zisserman, A.: Very deep convolutional networks for large-scale
  image recognition.
\newblock arXiv preprint arXiv:1409.1556  (2014)

\bibitem{googlenet}
Szegedy, C., Liu, W., Jia, Y., Sermanet, P., Reed, S., Anguelov, D., Erhan, D.,
  Vanhoucke, V., Rabinovich, A.: Going deeper with convolutions.
\newblock In: Proceedings of the IEEE conference on computer vision and pattern
  recognition, pp. 1--9 (2015)

\bibitem{deepface}
Taigman, Y., Yang, M., Ranzato, M., Wolf, L.: Deepface: Closing the gap to
  human-level performance in face verification.
\newblock In: Proceedings of the IEEE conference on computer vision and pattern
  recognition, pp. 1701--1708 (2014)

\bibitem{tzeng2015simultaneous}
Tzeng, E., Hoffman, J., Darrell, T., Saenko, K.: Simultaneous deep transfer
  across domains and tasks.
\newblock In: Proceedings of the IEEE International Conference on Computer
  Vision, pp. 4068--4076 (2015)

\bibitem{matconvnet}
Vedaldi, A., Lenc, K.: Matconvnet: Convolutional neural networks for matlab.
\newblock In: Proceedings of the 23rd ACM international conference on
  Multimedia, pp. 689--692. ACM (2015)

\bibitem{viola}
Viola, P., Jones, M.: Rapid object detection using a boosted cascade of simple
  features.
\newblock In: Computer Vision and Pattern Recognition, 2001. CVPR 2001.
  Proceedings of the 2001 IEEE Computer Society Conference on, vol.~1, pp.
  I--I. IEEE (2001)

\bibitem{wan2014deep}
Wan, J., Wang, D., Hoi, S.C.H., Wu, P., Zhu, J., Zhang, Y., Li, J.: Deep
  learning for content-based image retrieval: A comprehensive study.
\newblock In: Proceedings of the 22nd ACM international conference on
  Multimedia, pp. 157--166. ACM (2014)

\bibitem{wang2019multi}
Wang, Y., Wang, G., Chen, C., Pan, Z.: Multi-scale dilated convolution of
  convolutional neural network for image denoising.
\newblock Multimedia Tools and Applications pp. 1--16 (2019)

\bibitem{wen2016discriminative}
Wen, Y., Zhang, K., Li, Z., Qiao, Y.: A discriminative feature learning
  approach for deep face recognition.
\newblock In: European Conference on Computer Vision, pp. 499--515. Springer
  (2016)

\bibitem{rrelu}
Xu, B., Wang, N., Chen, T., Li, M.: Empirical evaluation of rectified
  activations in convolutional network.
\newblock arXiv preprint arXiv:1505.00853  (2015)

\bibitem{polyunir}
Zhang, B., Zhang, L., Zhang, D., Shen, L.: Directional binary code with
  application to polyu near-infrared face database.
\newblock Pattern Recognition Letters \textbf{31}(14), 2337--2344 (2010)

\bibitem{zhou2019deep}
Zhou, H., Li, Z.: Deep networks with non-static activation function.
\newblock Multimedia Tools and Applications \textbf{78}(1), 197--211 (2019)

\end{thebibliography}
}

\end{document}